\documentclass[5p,times]{elsarticle}
\usepackage{amssymb}
\usepackage{amsmath}
\usepackage{graphicx}
\usepackage{epstopdf}
\usepackage{hhline,rotating,multirow}
\usepackage{algorithmic,algorithm}
\usepackage{todonotes}
\usepackage{enumitem}
\usepackage{makecell}

\usepackage{tabularx}
\usepackage{colortbl}

\xdefinecolor{gray95}{gray}{0.65}
\xdefinecolor{gray25}{gray}{0.8}

\newcolumntype{b}{X}
\newcolumntype{s}{>{\hsize=.5\hsize}X}

\usepackage{booktabs}
\usepackage{array}
\usepackage{hyperref}

\definecolor{ambar}{rgb}{0.99, 0.99, 0.59}
\usepackage{listings}
\definecolor{codegray}{rgb}{0.2,0.2,0.2}

\journal{journal (under review)}

\begin{document}

\begin{frontmatter}

\title{jMetalPy: a Python Framework for Multi-Objective Optimization with Metaheuristics}


\author[uma]{Antonio Ben\'itez-Hidalgo}
\ead{antonio.b@uma.es}
\author[uma]{Antonio J. Nebro}
\ead{antonio@lcc.uma.es}
\author[uma]{Jos\'e Garc\'ia-Nieto}
\ead{jnieto@lcc.uma.es}
\author[tec]{Izaskun Oregi}
\ead{izaskun.oregui@tecnalia.com}
\author[tec,upv,bcam]{Javier Del Ser}
\ead{javier.delser@tecnalia.com}

\address[uma]{Departamento de Lenguajes y Ciencias de la Computaci\'on, Ada Byron Research Building, University of M\'{a}laga, 29071 M\'{a}laga, Spain} 
\address[tec]{TECNALIA, 48160 Derio, Spain}
\address[upv]{University of the Basque Country (UPV/EHU), 48013 Bilbao, Spain}
\address[bcam]{Basque Center for Applied Mathematics (BCAM), 48009 Bilbao, Spain}


\begin{abstract}
This paper describes jMetalPy, an object-oriented Python-based framework for multi-objective optimization with metaheuristic techniques. Building upon our experiences with the well-known jMetal framework, we have developed a new multi-objective optimization software platform aiming not only at replicating the former one in a different programming language, but also at taking advantage of the full feature set of Python, including its facilities for fast prototyping and the large amount of available libraries for data processing, data analysis, data visualization, and high-performance computing. As a result, jMetalPy provides an environment for solving multi-objective optimization problems focused not only on traditional metaheuristics, but also on techniques supporting preference articulation and dynamic problems, along with a rich set of features related to the automatic generation of statistical data from the results generated, as well as the real-time and interactive visualization of the Pareto front approximations produced by the algorithms. jMetalPy offers additionally support for parallel computing in multicore and cluster systems. We include some use cases to explore the main features of jMetalPy and to illustrate how to work with it.
\end{abstract}


\end{frontmatter}

\section{Introduction}
\label{sec:introduction}
Multi-objective optimization problems are widely found in many disciplines~\cite{Coello2006,Deb01}, including engineering, economics, logistics, transportation or energy, among others. They are characterized by having two or more conflicting objective functions that have to be maximized or minimized at the same time, with their optimum composed by a set of trade-off solutions known as Pareto optimal set. Besides having several objectives, other factors can make this family of optimization problems particularly difficult to tackle and solve with exact techniques, such as deceptiveness, epistasis, NP-hard complexity, or high dimensionality~\cite{Weise09}. As a consequence, the most popular techniques to deal with complex multi-objective optimization problems are metaheuristics~\cite{BR2003}, a family of non-exact algorithms including evolutionary algorithms and swarm intelligence methods (e.g. ant colony optimization or particle swarm optimization).

An important factor that has ignited the widespread adoption of metaheuristics is the availability of software tools easing their implementation, execution and deployment in practical setups. In the context of multi-objective optimization, one of the most acknowledged frameworks is jMetal~\cite{Durillo2011}, a project started in 2006 that has been continuously evolving since then, including a full redesign from scratch in 2015~\cite{NDV15}. jMetal is implemented in Java under the MIT licence, and its source code is publicly available in GitHub\footnote{jMetal: {\tt https://github.com/jMetal/jMetal}. As of \today, the papers about jMetal had accumulated more than 1280 citations (source: Google Scholar)}.

In this paper, we present jMetalPy, a new multi-objective optimization framework written in Python. Our motivation for developing jMetalPy stems from our past experience with jMetal and from the fact that nowadays Python has become a very prominent programming language with a plethora of interesting features, which enables fast prototyping fueled by its large ecosystem of libraries for numerical and scientific computing (NumPy~\cite{Numpy}, Scipy~\cite{Scipy}), data analysis (Pandas), machine learning (Scikit-learn~\cite{scikit-learn}), visualization (Matplotlib~\cite{Hunter:2007}, Holoviews ~\cite{holo}, Plotly~\cite{plotly}), large-scale processing (Dask \cite{Dask}, PySpark \cite{salloum2016big}) and so forth. Our goal is not only to rewrite jMetal in Python, but to focus mainly on aspects where Python can help fill the gaps not covered by Java. In particular, we place our interest in the analysis of results provided by the optimization algorithms, real-time and interactive visualization, preference articulation for supporting decision making, and solving dynamic problems. Furthermore, since Python can be thought of as a more agile programming environment for prototyping new multi-objective solvers, jMetalPy also incorporates a full suite of statistical significance tests and related tools for the sake of a principled comparison among multi-objective metaheuristics. 

jMetalPy has been developed by Computer Science engineers and scientists to support research in multi-objective optimization with metaheuristics, and to utilize the provided algorithms for solving real-word problems. Following the same open source philosophy as in jMetal, jMetalPy is released under the MIT license. The project is in continuous development, with its source code hosted in GitHub\footnote{jMetalPy: \url{https://github.com/jMetal/jMetalPy}}, where the last stable and current development versions can be freely obtained.

The main features of jMetalPy are summarized as follows:
\begin{itemize}[leftmargin=*]
\item jMetalPy is implemented in Python (version 3.6+), and its object-oriented architecture makes it flexible and extensible.
\item It provides a set of classical multi-objective metaheuristics (NSGA-II~\cite{DPA02}, GDE3~\cite{Kukkonen2005}, SMPSO~\cite{Nebro2009}, OMOPSO~\cite{omopso2004}, MOEA/D~\cite{Qingfu2007}) and standard families of problems for benchmarking (ZDT, DTLZ, WFG~\cite{Deb01}, and LZ09~\cite{LZ09}). 
\item Dynamic multi-objective optimization is supported, including the implementation of dynamic versions of NSGA-II and SMPSO, as well as the FDA~\cite{FDA04} problem family.
\item Reference point based preference articulation algorithms, such as SMPSO/RP~\cite{smpsorp2018} and versions of NSGA-II and GDE3, are also provided.
\item It implements quality indicators for multi-objective optimization, such as Hypervolume~\cite{ZT99}, Additive Epsilon~\cite{ZTL+03} and Inverted Generational Distance~\cite{CR04}.

\item It provides visualization components to display the Pareto front approximations when solving problems with two objectives (scatter plot), three objectives (scatter plot 3D), and many-objective problems (parallel coordinates graph and a tailored version of Chord diagrams).
\item Support for comparative studies, including a wide number of statistical tests and utilities (e.g. non-parametric test, post-hoc tests, boxplots, CD plot), including the automatic generation of \LaTeX{} tables (mean, standard deviation, median, interquartile range) and figures in different formats. 
\item jMetalPy can cooperatively work alongside with jMetal. The latter can be used to run algorithms and compute the quality indicators, while the post-processing data analysis can be carried out with jMetalPy.
\item Parallel computing is supported based on Apache Spark~\cite{Zaharia+2010} and Dask~\cite{Dask}. This includes an evaluator component that can be used by generational metaheuristics to evaluate solutions in parallel with Spark  (synchronous parallelism), as well as a parallel version of NSGA-II based on Dask (asynchronous parallelism). 
\item Supporting documentation. A website\footnote{jMetalPy documentation: \url{https://jmetalpy.readthedocs.io}} is maintained with user manuals and API specification for developers. This site also contains a series of Jupyter notebooks\footnote{Jupyter: \url{https://jupyter.org}} with use cases and examples of experiments and visualizations.
\end{itemize}

Our purpose of this paper is to describe jMetalPy, and to illustrate how it can be used by members of the community interested in experimenting with metaheuristics for solving multi-objective optimization problems. To this end, we include some implementation use cases based on NSGA-II to explore the main variants considered in jMetalPy, from standard versions (generational and steady state), to dynamic, reference-point based, parallel and distributed flavors of this solver. A experimental use case is also described to exemplify how the statistical tests and visualization tools included in jMetalPy can be used for post-processing and analyzing the obtained results in depth. For background concepts and formal definitions of multi-objective optimization, we refer to our previous work in~\cite{Durillo2011}.

The remaining of this paper is organized as follows. In Section~\ref{sec:relatedworks}, a review of relevant related algorithmic software platforms is conducted to give an insight and rationale of the main differences and contribution of jMetalPy. Section~\ref{sec:architecture} delves into the jMetalPy architecture and its main components. Section~\ref{sec:impl-usecase} explains a use case of implementation. Visualization facilities are described in Section~\ref{sec:visualization}, while a use case of experimentation with statistical procedures is explained in Section~\ref{sec:experimentation}. Finally, Section~\ref{sec:conclusions} presents the conclusions and outlines further related work planned for the near future.


\section{Related Works}
\label{sec:relatedworks}

{
\setlength{\tabcolsep}{0pt}
\begin{table*}[!ht]
    \centering
    \footnotesize
    \renewcommand{\arraystretch}{1.2}
    \caption{Most popular optimization frameworks written in Python.}
    \begin{tabularx}{\linewidth}{l@{\hskip 0.1in}l@{\hskip 0.1in}l@{\hskip 0.1in}l@{\hskip 0.1in}c@{\hskip 0.1in}c@{\hskip 0.1in}c@{\hskip 0.1in}l@{\hskip 0.1in}l}
        \toprule
        \emph{Name}  & \emph{Status} & \emph{Python} & \emph{License} & \emph{Parallel} & \emph{Dynamic} & \emph{Decision} & \emph{Post-processing} & \emph{Algorithms} \\
                    &                & \emph{version} &              & \emph{processing} & \emph{optimization} & \emph{making} & \emph{facilities} & \\
        \hline \hline
        DEAP 1.2.2~\cite{DEAP_JMLR2012}   & Active   & $\geq$2.7          & LGPL-3.0 & \checkmark & & & Statistics  & {\scriptsize GA, GP, CMA-ES, NSGA-II, SPEA2, MO-CMA-ES} \\
        Geatpy 1.1.5~\cite{geatpy}                   & Active   & $\geq$3.5          & MIT      &  &   & &  & {\scriptsize GA, MOEA}\\
        Inspyred 1.0.1~\cite{inspyred}    & Inactive & $\geq$2.6          & MIT      &  &  & &  &  {\scriptsize GA, ES, PSO, ACO, SA, PAES, NSGA-II} \\
        PyGMO 2.10~\cite{pagmo}    & Active   & 3.x                & GPL-3.0  & \checkmark & & & \makecell[l]{Visualization,\\ statistics}  & {\scriptsize \makecell[l]{GA, DE, PSO, SA, ABC, IHS, MC,\\ CMA-ES, NSGA-II, MOEA/D}}\\
        Platypus 1.0.3~\cite{Platypus}    & Active   & 3.6                & GPL-3.0  & \checkmark & & & \makecell[l]{Visualization,\\ statistics} & {\scriptsize \makecell[l]{CMA-ES, NSGA-II, NSGA-III, \\ GDE3, IBEA, MOEA/D,\\ OMOPSO, EpsMOEA, SPEA2}} \\
        Pymoo 0.2.4~\cite{blank19}                       & Active   & 3.6                & Apache 2.0  & & & \checkmark & \makecell[l]{Visualization,\\ statistics} &  \makecell[l]{{\scriptsize GA, DE, NSGA-II, NSGA-III,} \\ {\scriptsize U-NSGA-III,} reference point ({\scriptsize R-NSGA-III})}\\
        \hline
        jMetalPy 1.0.0                   & Active   & $\geq$3.6                & MIT  & \checkmark  & \checkmark & \checkmark & \makecell[l]{Visualization,\\ statistics}   & \makecell[l]{{\scriptsize GA, EA, NSGA-II, NSGA-III,}\\ {\scriptsize SMPSO, GDE3, OMOPSO, MOEA/D,}\\ reference point ({\scriptsize G-NSGA-II, SMPSO/RP, G-GDE3}), \\ dynamic ({\scriptsize NSGA-II, SMPSO, GDE3})} \\
        \bottomrule
    \end{tabularx}
    \label{table:related-frameworks}
\end{table*}
}

In the last two decades, a number of software frameworks devoted to the implementation of multi-objective metaheuristics has been contributed to the community, such as ECJ~\cite{Luke2017}, EvA~\cite{WZ97}, JCLEC-MO~\cite{jclec15}, jMetal~\cite{Durillo2011, NDV15}, MOEA Framework \cite{MOEAFramework}, and Opt4J~\cite{opt4jpaper}, which are written in Java; ParadisEO-MOEO \cite{Liefooghe10a}, and PISA~\cite{Bleuler02}, developed in C/C++; and PlatEMO \cite{Platemo17}, implemented in Matlab. They all have in common the inclusion of representative algorithms from the the state of the art, benchmark problems and quality indicators for performance assessment.

As has been mentioned in the introduction, there is a growing interest within the scientific community in software frameworks implemented in Python, since this language offers a large ecosystem of libraries, most of them devoted to data analysis, data processing and visualization. When it comes to optimization algorithms, a set of representative Python frameworks is listed in Table~\ref{table:related-frameworks}, where they are analyzed according to their algorithmic domains, maintenance status, Python version and licensing, as well as the featured variants, post-processing facilities and algorithms they currently offer. With the exception of the Inspyred framework, they are all active projects (i.e., their public source code have been updated at least one time within the last six months) and work out-of-the-box with a simple {\it pip} command. All of these frameworks support Python 3.x. 

DEAP and Inspyred are not centered in multi-objective optimization, and they include a shorter number of implemented algorithms. Pagmo/PyGMO, Platypus and Pymoo offer a higher number of features and algorithmic variants, including methods for statistical post-processing and visualization of results. In particular, Pagmo/PyGMO contains implementations of a number of single/multi-objective algorithms, including hybrid variants, with statistical methods for racing algorithms, quality indicators and fitness landscape analysis. Platypus supports parallel processing in solution evaluation phase, whereas Pymoo is rather focused on offering methods for preference articulation based on reference points.

The jMetalPy framework we proposed in this paper is also an active open source project, which is focused mainly on multi-objective optimization (although a number of single-objective algorithms are included) providing an increasing number of algorithms and modern methods for statistical post-processing and visualization of results. It offers algorithmic variants with methods for parallel processing and preference articulation based on reference points to provide decision making support. Moreover, jMetalPy incorporates algorithms and mechanisms for dynamic problem optimization, which is an additional feature not present in the other related frameworks. In this way, the proposed framework attempts at covering as many enhancing features in optimization as possible to support experimentation and decision making in both research and industry communities. Besides these features, an important design goal in jMetalPy has been to make the code easy to understand (in particular, the implementation of the algorithms), to reuse and to extend, as is illustrated in the next two sections.

\section{Architecture of jMetalPy}
\label{sec:architecture}

\begin{figure*}[!t]
    \centering
    \includegraphics[width=0.8\linewidth]{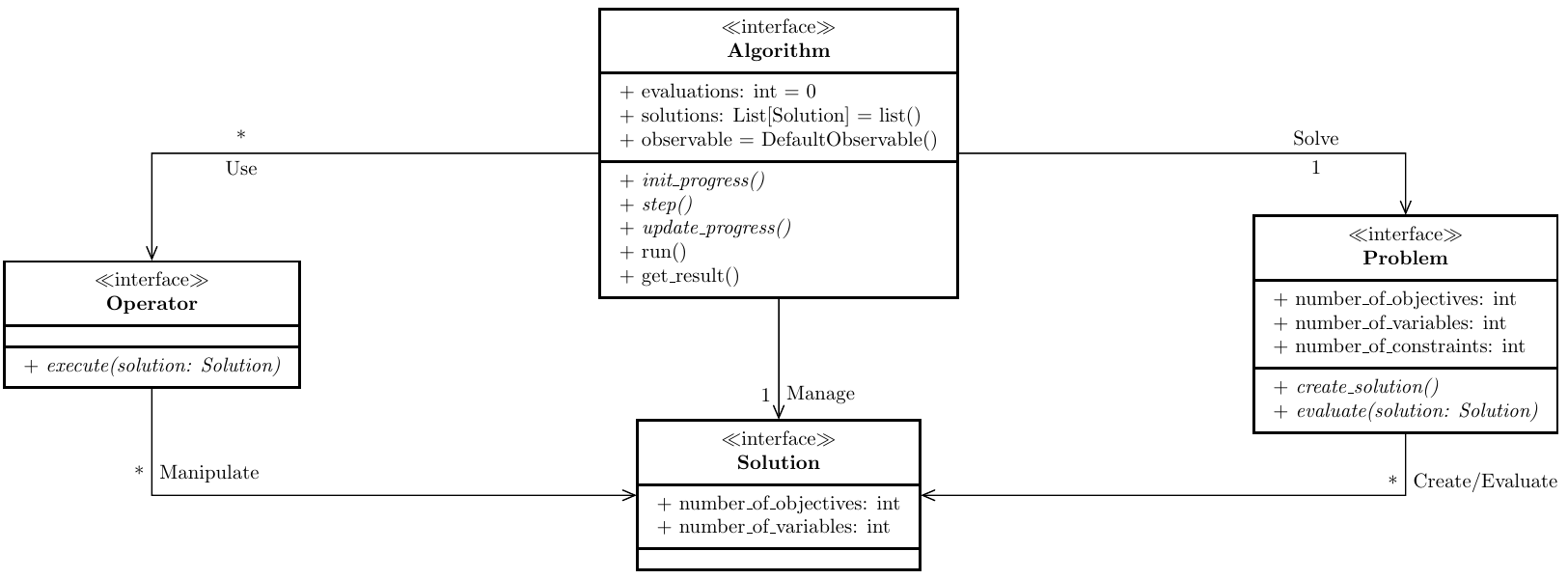}
    \caption{UML class diagram of jMetalPy.}
    \label{figure:uml-jmetalpy}
\end{figure*}

The architecture of jMetalPy has an object-oriented design to make it flexible and extensible (see Figure~\ref{figure:uml-jmetalpy}). The core classes define the basic functionality of jMetalPy: an {\it Algorithm} solves a {\it Problem} by using some {\it  Operator}  entities which manipulate a set of {\it Solution} objects. We detail these classes next.

\subsection{Core Architecture}
Class {\it Algorithm} contains a list of solutions (i.e. population in Evolutionary Algorithms or swarm in Swarm Intelligence techniques) and a {\it run()} method that implements the behavior of a generic metaheuristic (for the sake of simplicity, full details of the codes are omitted):
\begin{lstlisting}
class Algorithm(ABC):
  def __init__(self):
    self.evaluations = 0
    self.solutions = List[]
    self.observable = DefaultObservable()

  def run(self):
    self.solutions = self.create_initial_solutions()
    self.solutions = self.evaluate(self.solutions)
    self.init_progress()
    while not self.stopping_condition_is_met():
        self.step()
        self.update_progress()
\end{lstlisting}

In the above code we note the steps of creating the initial set of solutions, their evaluation, and the main loop of the algorithm, which performs a number of steps until a stopping condition is met. The initialization of state variables of an algorithm and their update at the end of each step are carried out in the {\it init\_progress()} and {\it update\_progress()} methods, respectively. In order to allow the communication of the status of an algorithm while running we have adopted the observer pattern~\cite{GHJ+94}, so that any algorithm is an observable entity which notifies to registered observers some information specified in advance (e.g., the current evaluation number, running time, or the current solution list), typically in the {\it update\_progress()} method. In this way we provide a structured method, for example, to display in real-time the current Pareto front approximation or to store it in a file. 

A problem is responsible of creating and evaluating solutions, and it is characterized by its number of decision variables, objectives and constraints. In case of the number of constraints be greater than 0, it is assumed that the {\it evaluate()} method also assesses whether the constraints are fulfilled. Subclasses of {\it Problem} include additional information depending of the assumed solution encoding; thus, a {\it FloatProblem} (for numerical optimization) or an {\it IntegerProblem} (for combinatorial optimization) requires the specification of the lower and upper bounds of the decision variables.

Operators such as {\it Mutation}, {\it Crossover}, and {\it Selection}, have an {\it execute(source)} method which, given a source object, produces a result. Mutations operate on a solution and return a new one resulting from modifying the original one. On the contrary, crossover operators take a list of solutions (namely, the parents) and produce another list of solutions (correspondingly, the offspring). Selection operators usually receive a list of solutions and returns one of them or a sublist of them.

The {\it Solution} class is a key component in jMetalPy because it is used to represent the available solution encodings, which are linked to the problem type and the operators that can be used to solve it. Every solution is composed by a list of variables, a list of objective values, and a set of attributes implemented as a dictionary of key-value pairs. Attributes can be used to assign, for example, a rank to the solutions of population or a constraint violation degree. Depending on the type of the variables, we have subclasses of {\it Solution} such as  {\it FloatSolution}, {\it IntegerSolution}, {\it BinarySolution} or {\it PermutationSolution}.

\subsection{Classes for Dynamic Optimization}

jMetalPy supports dealing with dynamic optimization problems, i.e., problems that change over time. For this purpose, it contains two abstract classes named {\it DynamicProblem} and {\it DynamicAlgorithm}.

A dynamic algorithm is defined as an algorithm with a restarting method, which is called whenever a change in the problem being solved is detected. The code of the {\it DynamicAlgorithm} class is as follows:
\begin{lstlisting}
class DynamicAlgorithm(Algorithm, ABC):

    @abstractmethod
    def restart(self) -> None:
        pass
\end{lstlisting}

The {\it DynamicProblem} class extends {\it Problem} with methods to query whether the problem has changed whatsoever, and to clear that status:
\begin{lstlisting}
class DynamicProblem(Problem, Observer, ABC):

    @abstractmethod
    def the_problem_has_changed(self) -> bool:
        pass

    @abstractmethod
    def clear_changed(self) -> None:
        pass
\end{lstlisting}

It is worth mentioning that a dynamic problem is also an observer entity according to the observer pattern. The underlying idea is that in jMetalPy it is assumed that changes in a dynamic problem are produced by external entities, i.e, observable objects where the problem is registered.  


\section{Implementation Use Case: NSGA-II and Variants}
\label{sec:impl-usecase}

With the aim of illustrating the basic usages of jMetalPy, in this section we describe the implementation of the well-known NSGA-II algorithm~\cite{DPA02}, as well as some of its variants (steady-state, dynamic, with preference articulation, parallel, and distributed). 

NSGA-II is a genetic algorithm, which is a subclass of Evolutionary Algorithms. In jMetalPy we include an abstract class for the latter, and a default implementation for the former. An Evolutionary Algorithm is a metaheuristic where the {\it step()} method consists of applying a sequence of selection, reproduction, and replacement methods, as illustrated in the code snippet below:
\begin{lstlisting}
class EvolutionaryAlgorithm(Algorithm, ABC):
  def __init__(self,
        problem: Problem,
        population_size: int,
        offspring_size: int):
    super(EvolutionaryAlgorithm, self).__init__()
    self.problem = problem
    self.population_size = population_size
    self.offspring_size = offspring_size

  @abstractmethod
  def selection(self, population):
    pass

  @abstractmethod
  def reproduction(self, population):
    pass

  @abstractmethod
  def replacement(self, population, offspring):
    pass

  def init_progress(self):
    self.evaluations = self.population_size

  def step(self):
    mating_pool = self.selection(self.solutions)
    offspring = self.reproduction(mating_pool)
    offspring = self.evaluate(offspring)
    self.solutions = self.replacement(self.solutions, offspring)

  def update_progress(self):
    self.evaluations += self.offspring_size
\end{lstlisting}

On every step, the selection operator is used (line 27) to retrieve the mating pool from the solution list (the population) of the algorithm. Solutions of the mating pool are taken for reproduction (line 28), which yields a new list of solutions called offspring. Solutions of this offspring population must be evaluated (line 29), and thereafter a replacement strategy is applied to update the population (line 30). We can observe that the evaluation counter is initialized and updated in the {\it init\_progress()} (line 23) and {\it update\_progress} (line 32), respectively.

The {\it EvolutionaryAlgorithm} class is very generic. We provide a complete implementation of a Genetic Algorithm, which is an evolutionary algorithm where the reproduction is composed by combining a crossover and mutation operator. We partially illustrate this implementation next:
\begin{lstlisting}
class GeneticAlgorithm(EvolutionaryAlgorithm):
  def __init__(self, 
        problem: Problem[Solution],
        population_size: int,
        offspring_population_size: int,
        mutation: Mutation,
        crossover: Crossover,
        selection: Selection,
        termination_criterion: TerminationCriterion,
        population_generator=RandomGenerator(),
        population_evaluator=SequentialEvaluator()):
    ...
	       
  def create_initial_solutions(self):
    return [self.population_generator.new(self.problem) 
            for _ in range(self.population_size)]

  def evaluate(self, solutions):
    return self.population_evaluator.evaluate(solutions, self.problem)
          
  def stopping_condition_is_met(self):
    return self.termination_criterion.is_met
         
  def selection(self, population: List[Solution]):
    # select solutions to get the mating pool
                
  def reproduction(self, mating_pool):
    # apply crossover and mutation
        
  def replacement(self, population, offspring):
    # combine the population and offspring populations 
\end{lstlisting}

There are some interesting features to point out here. First, the initial solution list is created from a {\it Generator} object (line 14), which, given a problem, returns a number of new solutions according to some strategy implemented in the generator; by default, a {\it RandomGenerator()} is chosen to produce a number of solutions uniformly drawn at random from the value range specified for the decision variables. Second, an {\it Evaluator} object is used to evaluate all produced solutions (line 19); the default one evaluates the solutions sequentially. Third, a {\it TerminationCriterion} object is used to check the stopping condition (line 21), which allows deciding among several stopping criteria when configured. The provided implementations include: stopping after making a maximum number of evaluations, computing for a maximum time, a key has been pressed, or the current population achieves a minimum level of quality according to some indicator.  Fourth, the reproduction method applies the crossover and mutation operators over the mating pool to generate the offspring population. Finally, the replacement method combines the population and the offspring population to produce a new population. 

Departing from the implemented {\it GeneticAlgorithm} class, we are ready to implement the standard NSGA-II algorithm and some variants, which will be described in the next subsections. Computing times will be reported when running the algorithm to solve the ZDT1 benchmark problem~\cite{zdt2000a} on a MacBook Pro with macOS Mojave, 2.2 GHz Intel Core i7 processor (Turbo boost up to 3.4GHz), 16 GB 1600 MHz DDR3 RAM, Python 3.6.7 :: Anaconda.

\subsection{Standard Generational NSGA-II}
\label{subsection:nsgaii}

\begin{figure}[!hb]
\centering
  \fbox{\includegraphics[width=0.9\columnwidth]{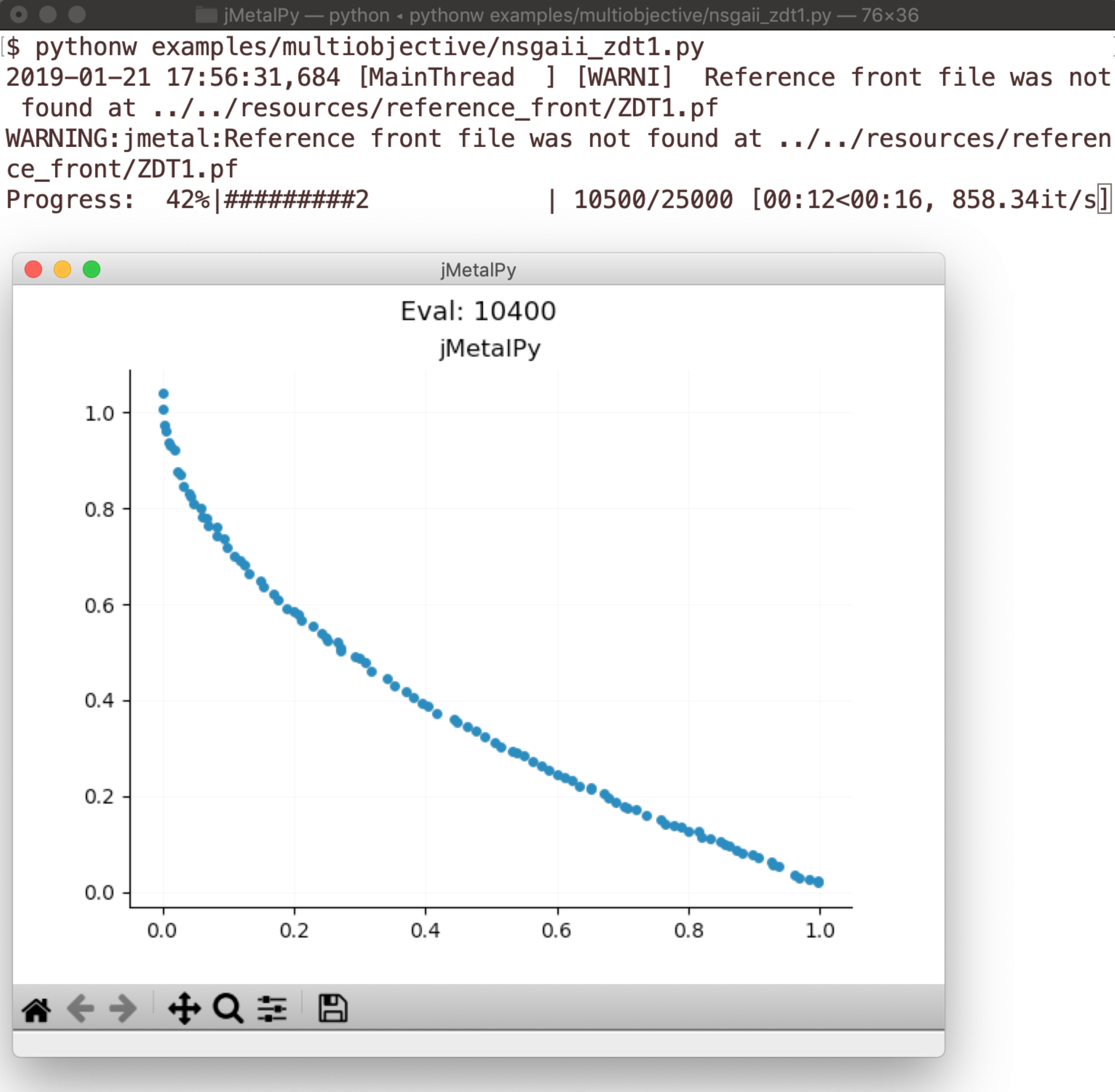}}
  \caption{Screenshot of jMetalPy running a NSGA-II for the ZDT1 benchmark problem showing the progress and the Pareto front approximation.}
  \label{figure:capture}
\end{figure}

\begin{figure*}
\centering
  \includegraphics[width=60mm]{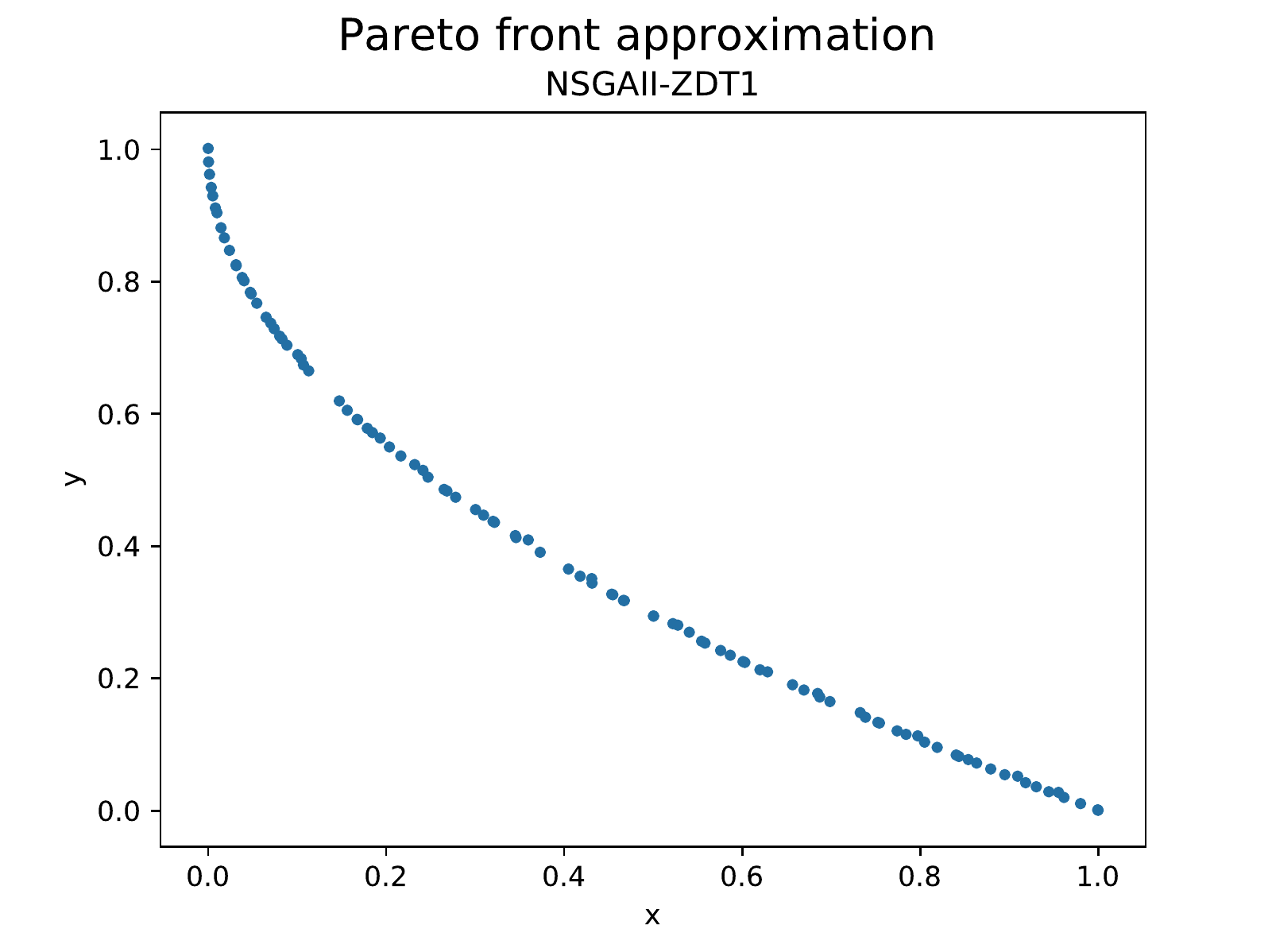}
  \includegraphics[width=60mm]{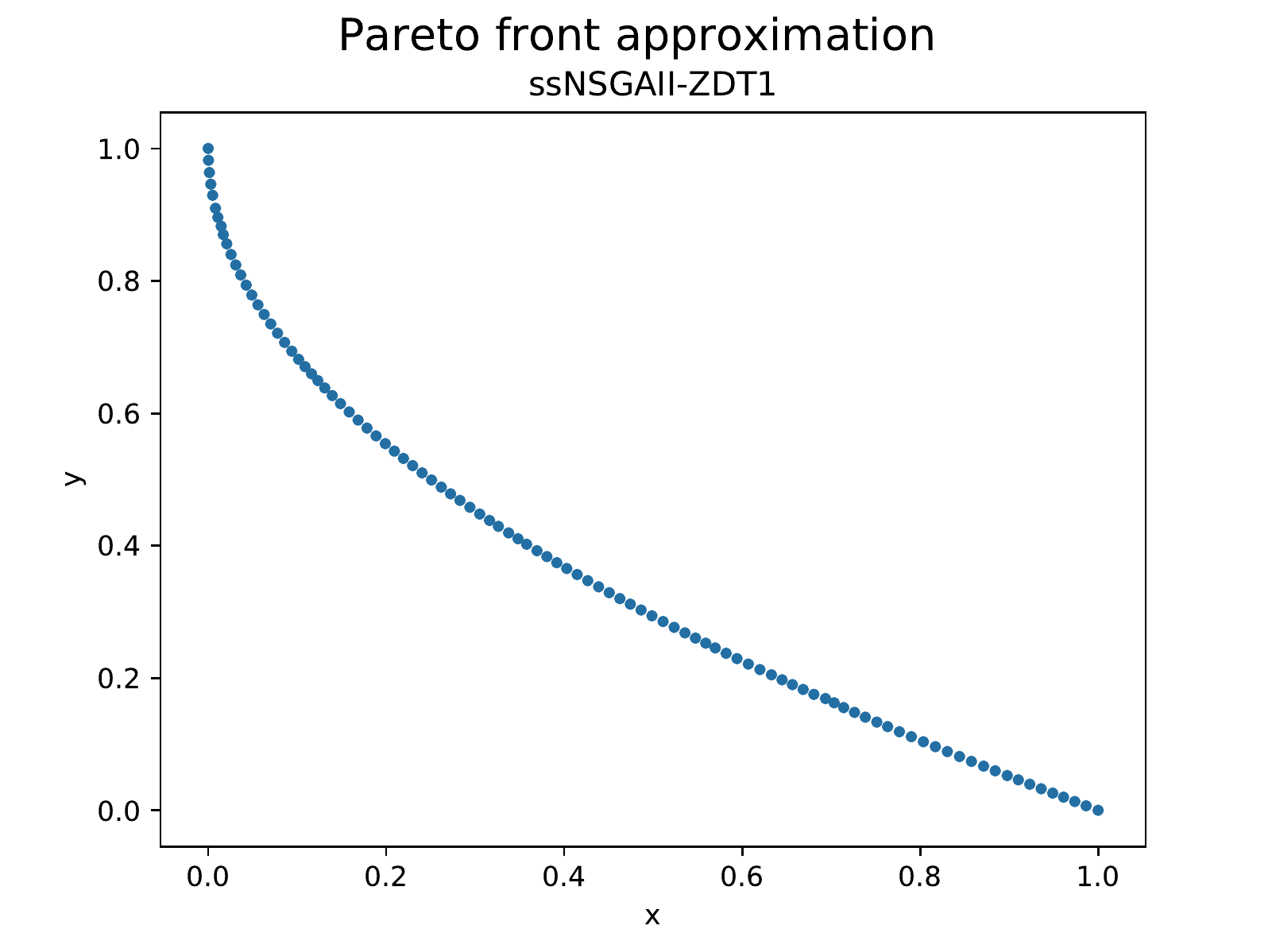}
  \includegraphics[width=60mm]{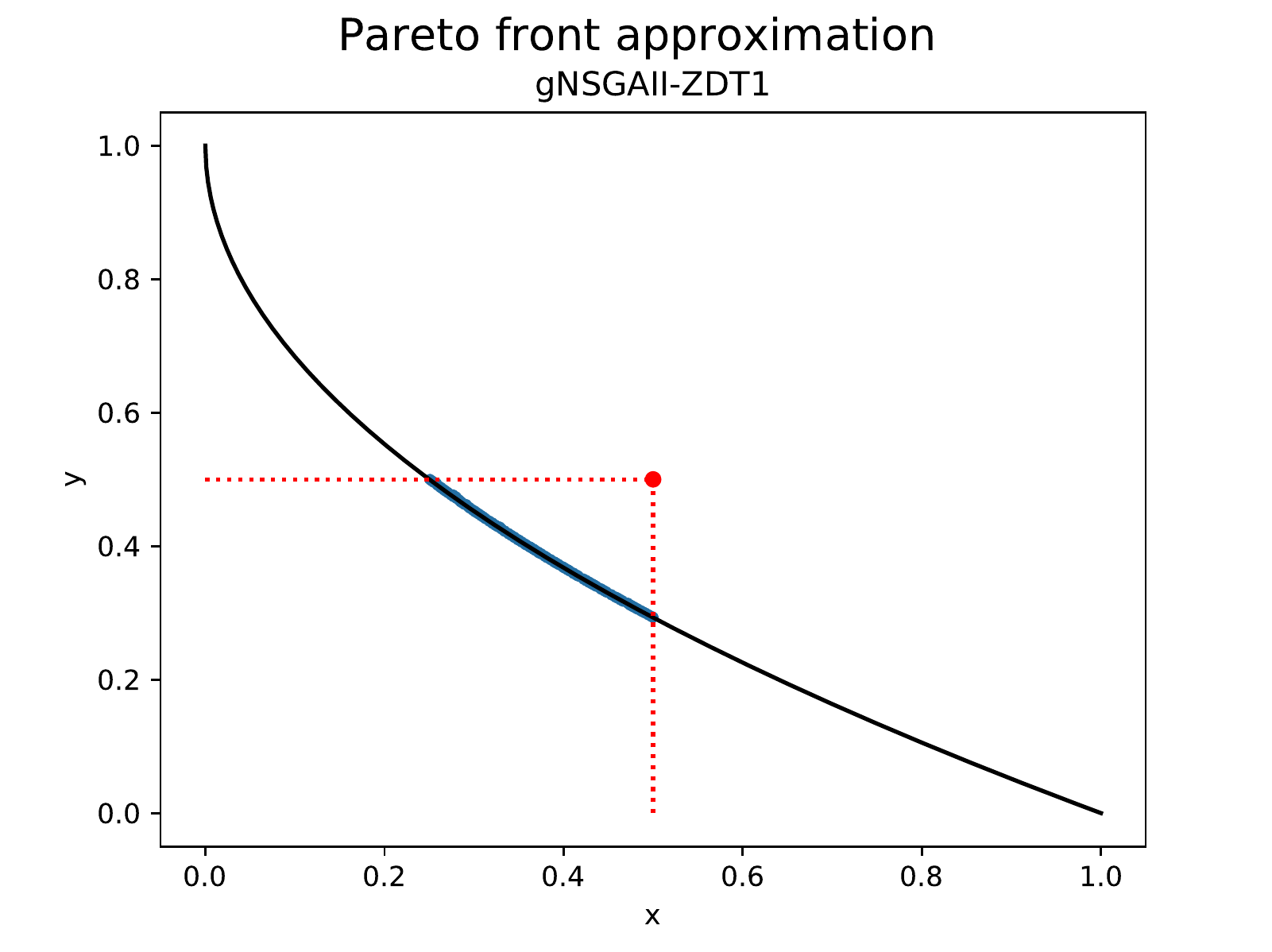}
  \caption{Pareto front approximations when solving the ZDT1 produced by the standard NSGA-II algorithm (left), a steady-state version (center), and G-NSGA-II (using the reference point $[f_1,f_2]=[0.5, 0.5]$, shown in red).}
  \label{figure:nsgaiifronts}
\end{figure*}

NSGA-II is a generational genetic algorithm, so the population and the offspring population have the same size. Its main feature is the use of a non-dominated sorting for ranking the solutions in a population to foster convergence, and a crowding distance density estimator to promote diversity~\cite{DPA02}. These mechanisms are applied in the replacement method, as shown in the following snippet:
\begin{lstlisting}
class NSGAII(GeneticAlgorithm):
  def __init__(self,
        problem: Problem,
        population_size,
        offspring_size,
        mutation: Mutation,
        crossover: Crossover,
        selection: Selection,
        termination_criterion: TerminationCriterion,
        population_generator=RandomGenerator(),
        population_evaluator=SequentialEvaluator()
        dominance_comparator=DominanceComparator()):
    ...
  def replacement(self, population, offspring):
    join_population = population + offspring
        
    return RankingAndCrowdingDistanceSelection(
        self.population_size, self.dominance_comparator).execute(join_population)
\end{lstlisting}

No more code is needed. To configure and run the algorithm we include some examples, such as the following code:
\begin{lstlisting}
# Standard generational NSGAII runner
problem = ZDT1()

max_evaluations = 25000
algorithm = NSGAII(
  problem=problem,
  population_size=100,
  offspring_population_size=100,
  mutation=PolynomialMutation(...),
  crossover=SBXCrossover(...),
  selection=BinaryTournamentSelection(...),
  termination_criterion=StoppingByEvaluations(max=max_evaluations),
  dominance_comparator=DominanceComparator()
)

progress_bar = ProgressBarObserver(max=max_evals)
algorithm.observable.register(observer=progress_bar)
  
real_time = VisualizerObserver()
algorithm.observable.register(observer=real_time)

algorithm.run()
front = algorithm.get_result()

# Save results to file
print_function_values_to_file(front, `FUN')
print_variables_to_file(front, `VAR')
\end{lstlisting}


This code snippet depicts a standard configuration of NSGA-II to solve the ZDT1 benchmark problem. Note that we can define a dominance comparator (line 13), which by default is the one used in the standard implementation of NSGA-II.

As commented previously, any algorithm is an observable entity, so observers can register into it. In this code, we register a progress bar observer (shows a bar in the terminal indicating the progress of the algorithm) and a visualizer observer (shows a graph plotting the current population, i.e., the current Pareto front approximation). A screen capture of NSGA-II running in included in Figure~\ref{figure:capture}. The computing time of NSGA-II with this configuration in our target laptop is around 9.2 seconds.

\subsection{Steady-State NSGA-II} \label{subsection:steadystatensgaii}

A steady-state version of NSGA-II can be configured by resorting to the same code, but just setting the offspring population size to one. This version yielded a better performance in terms of the produced Pareto front approximation compared with the standard NSGA-II as reported in a previous study~\cite{DNL08}, but at a cost of a higher computing time, which raises up to 190 seconds. 

An example of Pareto front approximation found by this version of NSGA-II when solving the ZDT1 benchmark problem is shown in Figure~\ref{figure:nsgaiifronts}-center. As expected given the literature, it compares favorably against the one generated by the standard NSGA-II (Figure~\ref{figure:nsgaiifronts}-left).

\subsection{NSGA-II with Preference Articulation} \label{subsection:preferencearticulationnsgaii}

The NSGA-II implementation in jMetalPy can be easily extended to incorporate a preference articulation scheme. Concretely, we have developed a g-dominance based comparator considering the {\textsc g}-dominance concept described in \cite{MOLINA2009685}, where a region of interest can be delimited by defining a reference point. If we desire to focus the search in the interest region delimited by the reference point, say e.g. $[f_1,f_2]=[0.5, 0.5]$,  we can configure NSGA-II with this comparator as follows:


\begin{lstlisting}
reference_point = [0.5, 0.5]
algorithm = NSGAII(
  ...
  dominance_comparator=GDominanceComparator(reference_point)
)
\end{lstlisting}

The resulting front is show in Figure~\ref{figure:nsgaiifronts}-right.

\subsection{Dynamic NSGA-II} \label{subsection:dynamicnsgaii}

The approach adopted in jMetalPy to provide support for dynamic problem solving is as follows: First, we have developed a {\it TimeCounter} class (which is an {\it Observable} entity) which, given a delay, increments continuously a counter and notifies the registered observers the new counter values; second, we need to define an instance of {\it DynamicProblem}, which must implement the methods for checking whether the problem has changed and to clear the changed state. As {\it DynamicProblem} inherits from {\it Observer}, instances of this class can register in a {\it TimeCounter} object. Finally, it is required  to extend {\it DynamicAlgorithm} with a class defining the {\it restart()} method that will be called when the algorithm detects a change in a dynamic problem. The following code snippet shows the implementation of the {\it DynamicNSGAII} class:
\begin{lstlisting}
class DynamicNSGAII(NSGAII, DynamicAlgorithm):
  def __init__(self, ...):
    ...
    self.completed_iterations = 0

  def restart(self) -> None
    # restart strategy

  def update_progress(self):
    if self.problem.the_problem_has_changed():
      self.restart()
      self.evaluator.evaluate(self.solutions, problem)
      self.problem.clear_changed()
      self.evaluations += self.offspring_size

  def stopping_condition_is_met(self):
    if self.termination_criterion.is_met:
      self.restart()
      self.evaluator.evaluate(self.solutions, problem)
      self.init_progress()
      self.completed_iterations += 1
\end{lstlisting}

As shown above, at the end of each iteration a check is made about a change in the problem. If a change has occurred, the restart method is invoked which, depending on the implemented strategy, will remove some solutions from the population and new ones will be created to replace them. The resulting population will be evaluated and the {\it clear\_changed()} method of the problem object will be called. As opposed to the standard NSGA-II, the stopping condition method is not invoked to halt the algorithm, but instead to notify registered observers (e.g., a visualizer) that a new resulting population has been produced. Then, the algorithm starts again by invoking the {\it restart()} and {\it init\_progress()} methods. 
It is worth noting that most of the code of the original NSGA-II implementation is reused and only some methods need to be rewritten. 

To illustrate the implementation a dynamic problem, we next show code of the {\it FDA} abstract class, which is the base class of the five problems composing the FDA benchmark:
\begin{lstlisting}
class FDA(DynamicProblem, FloatProblem, ABC):
    def __init__(self):
        super(FDA, self).__init__()
        self.tau_T = 5
        self.nT = 10
        self.time = 1.0
        self.problem_modified = False

    def update(self, *args, **kwargs):
        counter = kwargs['COUNTER']
        self.time = (1.0 / self.nT) * floor(counter * 1.0 / self.tau_T)
        self.problem_modified = True

    def the_problem_has_changed(self) -> bool:
        return self.problem_modified

    def clear_changed(self) -> None:
        self.problem_modified = False
\end{lstlisting}

The key point in this class is the {\it update()} method which, when invoked by an observable entity (e.g., an instance of the aforementioned {\it TimeCounter} class), sets the {\it problem\_modified} flag to True. We can observe that this flag can be queried and reset.

\begin{figure}[!h]
\centering
  \includegraphics[width=80mm]{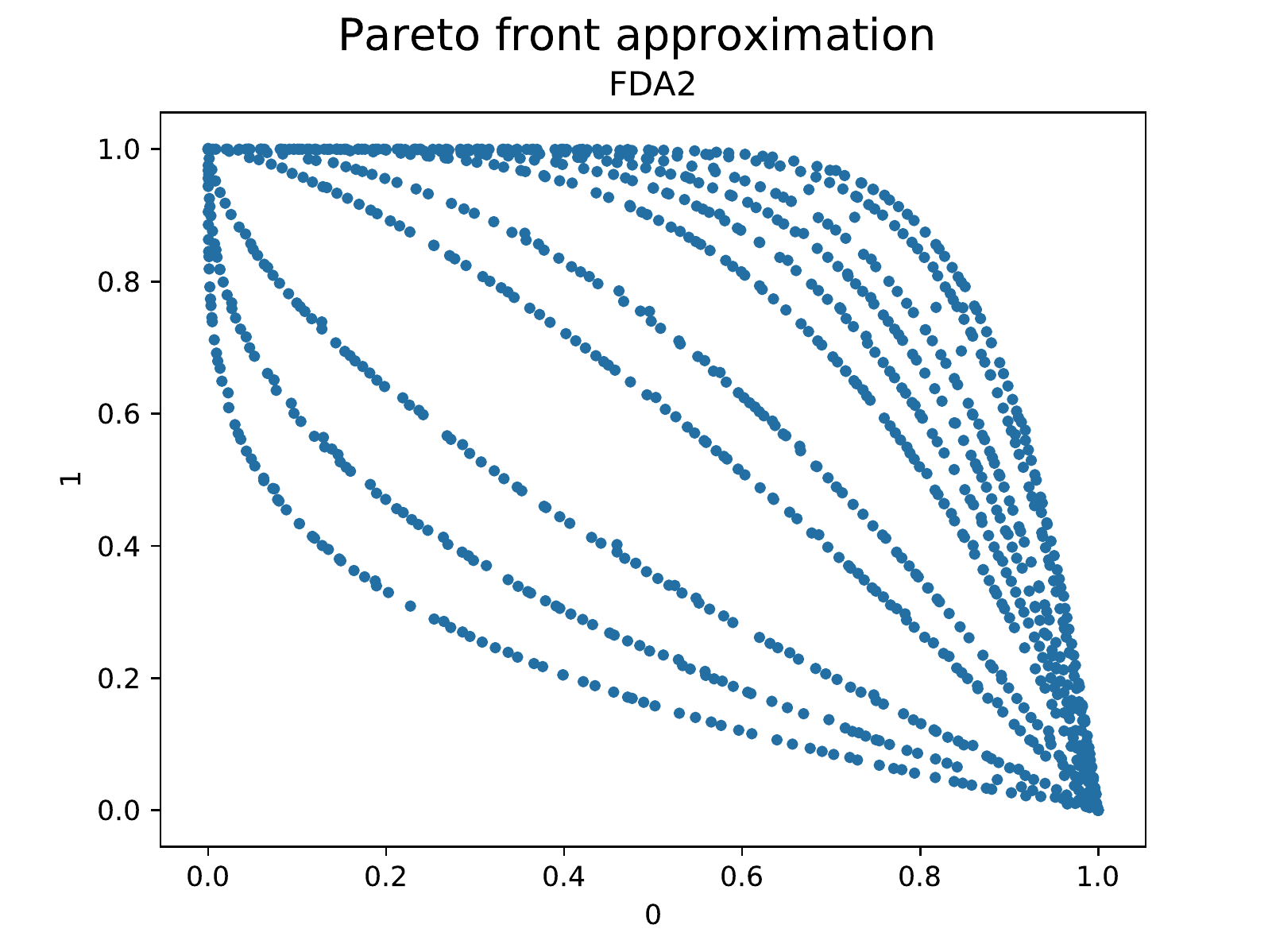}
  \caption{Pareto front approximations when solving the dynamic FDA2 problem produced by the dynamic version of NSGA-II.}
  \label{figure:dynamicnsgaii}
\end{figure}

The code presented next shows how to configure and run the dynamic NSGA-II algorithm:

\begin{lstlisting}
# Dynamic NSGAII runner
problem = FDA2()
time_counter = TimeCounter(delay=1))
time_counter.observable.register(problem)
time_counter.start()

algorithm = DynamicNSGAII(
  ...
  termination_criterion=StoppingByEvaluations(max=
  max_evals)
  )
algorithm.run()
\end{lstlisting}

After creating the instances of the FDA2 benchmark problem~\cite{FDA04} and the time counter class, the former is registered in the latter, which runs in a concurrent thread. The dynamic NSGA-II is set with stopping condition which returns a Pareto front approximation every 25,000 function evaluations. An example of running of the dynamic NSGA-II algorithm when solving the FDA2 problem is shown in Figure~\ref{figure:dynamicnsgaii}.

\subsection{Parallel NSGA-II with Apache Spark} \label{subsection:parallelnsgaiispak}

In order to evaluate a population, NSGA-II (and in general, any generational algorithms in jMetalPy) can use an evaluator object. The default evaluator runs in a sequential fashion but, should the evaluate method of the problem be thread-safe, solutions can be evaluated in parallel. jMetalPy includes an evaluator based on Apache Spark, so the solutions can be evaluated in a variety of parallel systems (multicores, clusters) following the scheme presented in~\cite{BGN17}. This evaluator can be used as exemplified next:
\begin{lstlisting}
# NSGAII runner using the Spark evaluator
algorithm = NSGAII(
  ...
  evaluator=SparkEvaluator()
)
\end{lstlisting}

The resulting parallel NSGA-II algoritm combines parallel with sequential phases, so speed improvements cannot be expected to scale linearly. A pilot test on our target laptop indicates speedup factors in the order of 2.7. However, what is interesting to note here is that no changes are required in NSGA-II, which has the same behavior as its sequential version, so the obtained time reductions are for free.

\subsection{Distributed NSGA-II with Dask}
\label{subsection:distributednsgaiidask}

The last variant of NSGA-II we present in this paper is a distributed version based on an asynchronous parallel model implemented with Dask~\cite{Dask}, a parallel and distributed Python system including a broad set of parallel programming models, including asynchronous parallelism using futures.

The distributed NSGA-II adopts a parallel scheme studied in~\cite{DNL08}. The scheme is based on a steady-state NSGA-II and the use of Dask's futures, in such a way that whenever a new solution has to evaluated, a task is created and submitted to Dask, which returns a future. When a task is completed, its corresponding future returns an evaluated solution, which is inserted into the offspring population. Then, a new solution is produced after performing the replacement, selection, and reproduction stages, to be sent again for evaluation. This way, all the processors/cores of the target cluster will be busy most of the time.

Preliminary results on our target multicore laptop indicate that speedups around 5.45 can obtained with the 8 cores of the system where simulations were performed. We will discuss on this lack of scalability and other aspects of this use case in the next subsection.


\subsection{Discussion}

In this section we have presented five different versions of NSGA-II , most of them (except for the distributed variant) requiring minor changes on the base class implementing NSGA-II. Not all algorithms can be adapted in the same way, but some of the variations of NSGA-II can be implemented in a straightforward manner. Thus, we include in jMetalPy examples of dynamic, preference-based, and parallel versions of some of the included algorithms, such as SMPSO, GDE3, and OMOPSO.

We would like to again stress on the readability of the codes, by virtue of which all the steps of the algorithms can be clearly identified. Some users may find the class hierarchy {\it EvolutionaryAlgorithm} $\rightarrow$ {\it GeneticAlgorithm} $\rightarrow$ {\it NSGAII} cumbersome, and prefer to have all the code of NSGA-II in a single class. However, this alternative design approach would hinder the flexibility of the current implementation, and would require to replicate most of the code when developing algorithmic variants.  

In the case of parallel algorithms, an exhaustive performance assessment is beyond the scope of this paper. The reported speedups are not remarkable due to the Turbo Boost feature of the processor of the laptop used for performing the experiments, but they give an idea of the time reductions that can be achieved when using a modern multicore computer. 

\section{Visualization}
\label{sec:visualization}

An advantage of using Python (instead of Java) is its power related to visualization features thanks to the availability of graphic plotting libraries, such as: Matplotlib, Holoviews or Plotly.

jMetalPy harnesses these libraries to include three types of visualization charts: static, interactive and streaming. Table~\ref{table:graphs} summarizes these implementations. Static charts can be shown in the screen, stored in a file, or included in a Jupyter notebook (typically used at the end of the execution of an algorithm). Similarly, interactive charts are generated when an algorithm returns a Pareto front approximation but, unlike the static ones, the user can manipulate them interactively. There are two kinds of interactive charts: those that produce an HTML page including a chart (allowing to apply actions such as zooming, selecting part of the graph, or clicking in a point to see its objective values are allowed) and charts such as the Chord diagram that allows hovering the mouse over the chart and visualizing relationships among objective values. Finally, streaming charts depict graphs in real time, during the execution of the algorithms (and they can also be included in a Jupyter notebook); this can be useful to observe the evolution of the current Pareto front approximation produced by the algorithm.
\begin{table}[!htbp]
    \centering
    \footnotesize
    \caption{Main visualizations included in jMetalPy.}
    \begin{tabularx}{\linewidth}{s@{\hspace{0.2cm}}ssb}
        \toprule
        \emph{Name} & \emph{Type} & \emph{Backend} & \emph{Description} \\
        \midrule
        Plot & Static & Matplotlib & 2D, 3D, p-coords \\
             & Interactive & Plotly & 2D, 3D, p-coords \\
        Streaming plot & Streaming & Matplotlib & 2D, 3D \\
                       & Streaming & HoloViews & 2D, 3D (for Jupyter) \\
        Chord plot & Interactive & Matplotlib & For statistical purposes\\
        Box plot & Interactive & Matplotlib & For statistical purposes\\
        CD plot & Static & Matplotlib & Demsar's critical distance plot \\
        Posterior plot & Static & Matplotlib & Bayesian posterior analysis \\
        \bottomrule
    \end{tabularx}
    \label{table:graphs}
\end{table}

\begin{figure}[!ht]
    \centering
    \includegraphics[width=0.85\columnwidth]{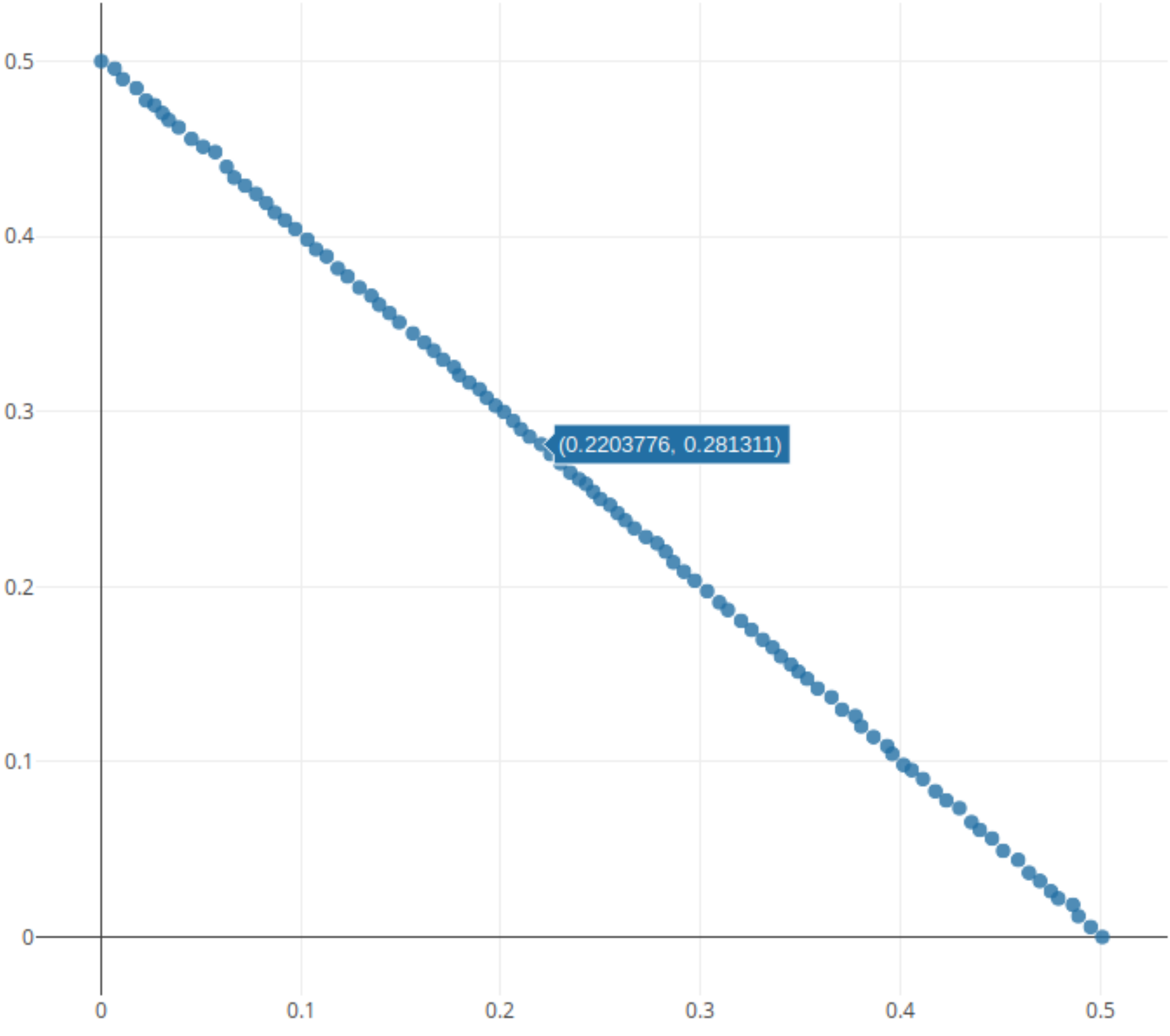}\vspace{0.4cm}
    \includegraphics[width=0.85\columnwidth]{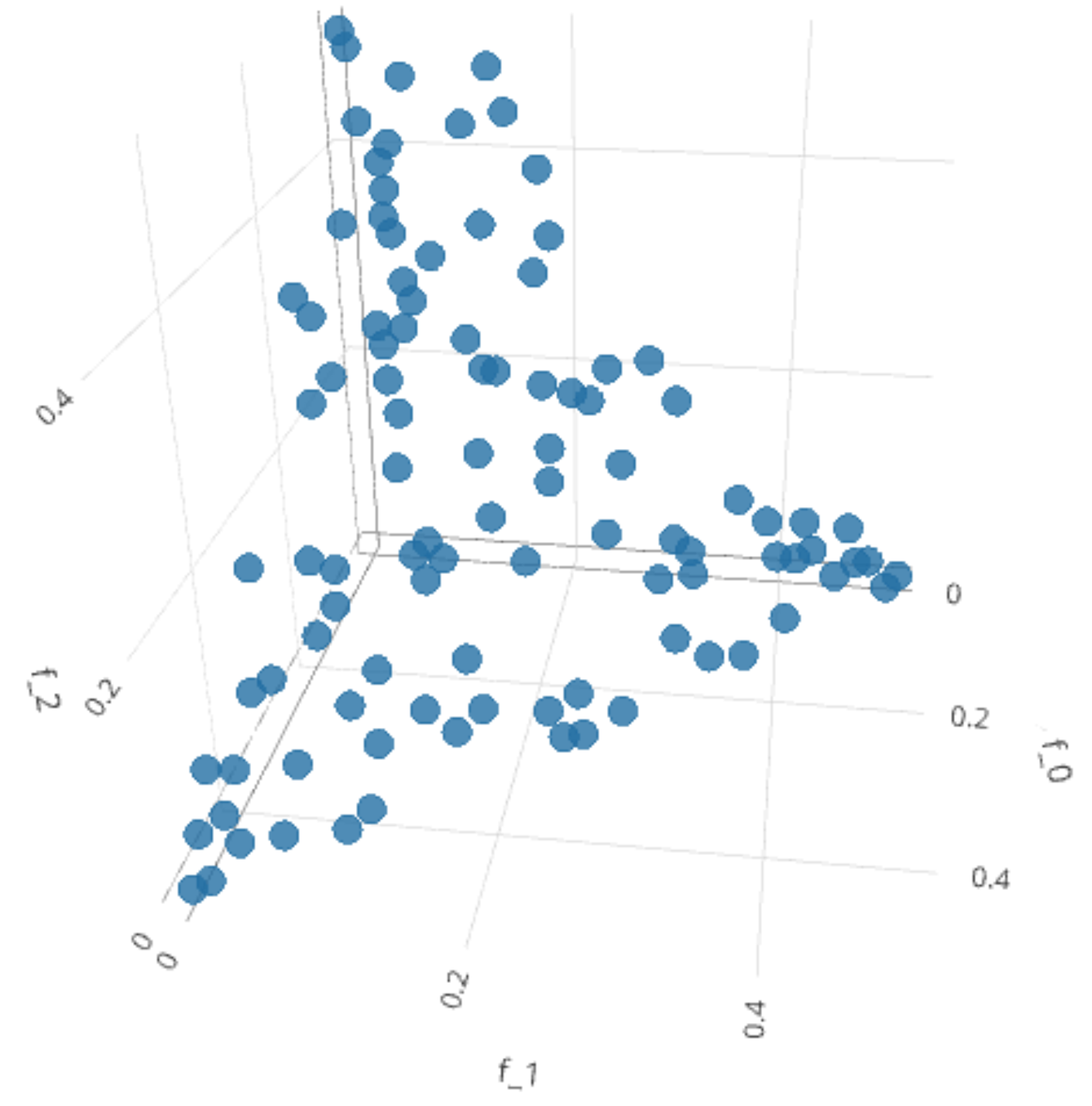}\vspace{0.4cm}
    \includegraphics[width=0.85\columnwidth]{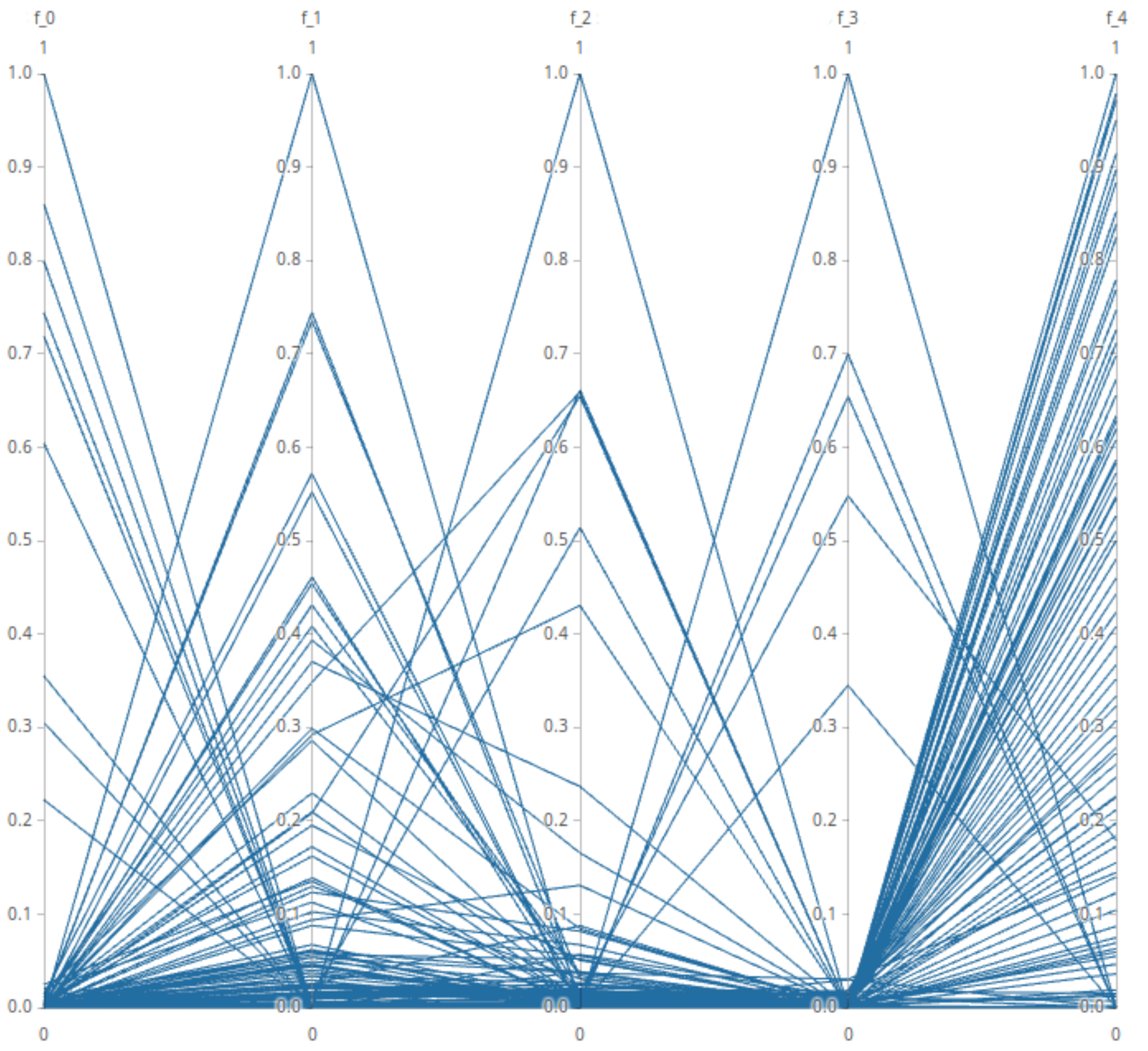}
    \caption{Examples of interactive plots produced when using SMPSO to solve the DTLZ1 problem with 2 (top), 3 (middle), and 5 (bottom) objectives.}
    \label{figure:interactiveplots}
\end{figure}

\begin{figure}
\centering
 \includegraphics[width=0.9\columnwidth]{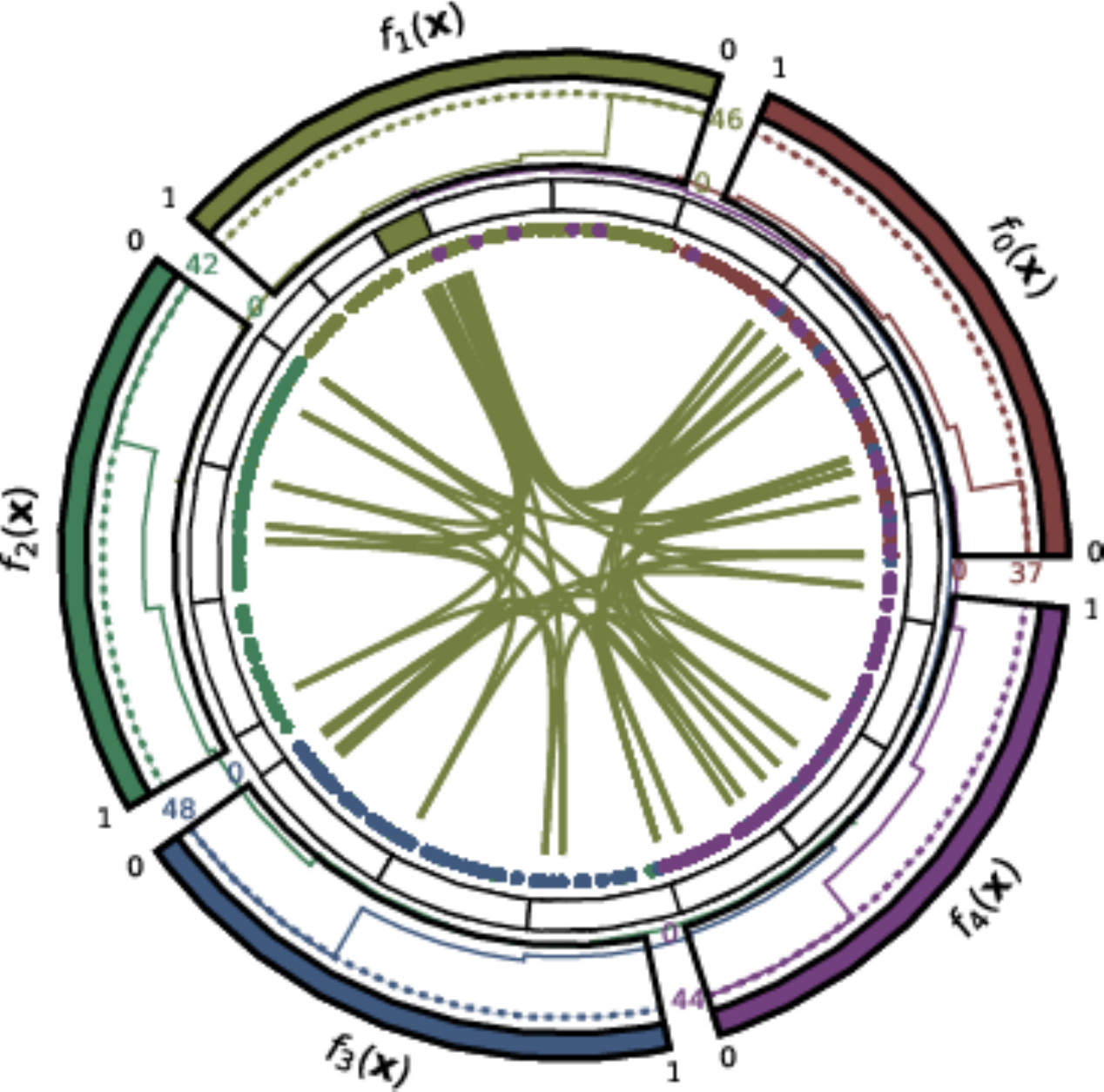}
  \caption{Example of Chord diagram for the front obtained by SMPSO when solving a problem with 5 objectives.}
  \label{figure:chorddiagram}
\end{figure}

Figure~\ref{figure:interactiveplots} shows three examples of interactive plots based on Plotly. The target problem is DTLZ1~\cite{DTLZ01}, which is solved with the SMPSO algorithm when the problem is defined with 2, 3 and 5 objectives. For any problem with more than 3 objectives, a parallel coordinates graph is generated. An example of Chord diagram for a problem with 5 objectives is shown in Figure~\ref{figure:chorddiagram}; each depicted chord represents a solution of the obtained Pareto front, and ties together its objective values. When hovering over a sector box of a certain objective $f_i$, this chart only renders those solutions whose $f_i$ values fall within the value support of this objective delimited by the extremes of the sector box. Finally, the outer partitioned torus of the chart represents a histogram of the values covered in the obtained Pareto front for every objective.

\section{Experimental Use Case} \label{sec:experimentation}

In previous sections, we have shown examples of Pareto front approximations produced by some of the metaheuristics included in jMetalPy. In this section, we describe how our framework can be used to carry out rigorous experimental studies based on comparing a number of algorithms to determine which of them presents the best overall performance. 

\subsection{Experimentation Methodology}
An experimental comparison requires a number of steps:
\begin{enumerate}[leftmargin=*]
    \item Determine the algorithms to be compared and the benchmark problems to be used.
    \item Run a number of independent runs per algorithm-problem configuration and get the produced fronts.
    \item Apply quality indicators to the fronts (e.g., Hypervolume, Epsilon, etc.).
    \item Apply a number of statistical test to assess the statistical significance of the performance differences found among the algorithms considered in the benchmark.
\end{enumerate}

The first three steps can be done with jMetalPy, but also with jMetal or even manually (e.g., running algorithms using a script). The point where jMetalPy stands out is the fourth one, as it contains a large amount of statistical features to provide the user with a broad set of tools to analyze the results generated by a comparative study. All these functionalities have been programmed from scratch and embedded into the core of jMetalPy. Specifically, the statistical tests included in jMetalPy are listed next:
\begin{itemize}[leftmargin=*]
    \item A diverse set of non-parametric null hypothesis significance tests, namely, the Wilcoxon rank sum test, Sign test, Friedman test, Friedman aligned rank test and Quade test. These tests have been traditionally used by the community to shed light on their comparative performance by inspecting a statistic computed from their scores.
    \item Bayesian tests (sign test and signed rank test), which have been recently postulated to overcome the shortcomings of null hypothesis significance testing for performance assessment \cite{benavoli2017time}. These tests are complemented by a posterior plot in barycentric coordinates to compare pairs of algorithms under a Bayesian approach by also accounting for possible statistical ties. 
    \item Posthoc tests to compare among multiple algorithms, either one-vs-all (Bonferroni-Dunn, Holland, Finner, and Hochberg) or all-vs-all (Li, Holm, Shaffer).
\end{itemize}

The results of these tests are displayed by default in the screen and most of them can be exported to \LaTeX{} tables. Furthermore, boxplot diagrams can be also generated. Finally, \LaTeX{} tables containing means and medians (and their corresponding standard deviation and interquartile range dispersion measures, respectively) are automatically generated.

\subsection{Implementation Details}
jMetalPy has a {\it laboratory} module containing utilities for defining experiments, which require three lists: the algorithms to be compared (which must be properly configured), the benchmark problems to be solved, and the quality indicators to be applied for performance assessment. Additional parameters are the number of independent runs and the output directory. 

Once the experiment is executed, a summary in the form of a CSV file is generated. This file contains all the information of the quality indicator values, for each configuration and run. Each line of this file has the following schema: Algorithm, Problem, Indicator, ExecutionId, IndicatorValue. An example of its contents follows:

\begin{lstlisting}
Algorithm,Problem,Indicator,ExecutionId,IndicatorValue
NSGAII,ZDT1,EP,0,0.015705992620067832
NSGAII,ZDT1,EP,1,0.012832504015918067
NSGAII,ZDT1,EP,2,0.01071189935186434
...
MOCell,ZDT6,IGD+,22,0.0047265135903854704
MOCell,ZDT6,IGD+,23,0.004496215669027173
MOCell,ZDT6,IGD+,24,0.005483899232523609
\end{lstlisting}

\noindent where we can see the header with the column names, followed by four lines corresponding to the values of the Epsilon indicator of three runs of the NSGA-II algorithm when solving the ZDT1 problem. The end of the file shows the value of the IGD+ indicator for three runs of MOCell when solving the ZDT6 problem. The file contains as many lines as the product of the numbers of algorithms, problems, quality indicators, and independent runs.

The summary file is the input of all the statistical tests, so that they can be applied to any valid file having the proper format. This is particularly interesting to combine  jMetal and jMetalPy. The last versions of jMetal generates a summary file after running a set of algorithms in an experimental study, so then we can take advantage of the features of jMetal (providing many algorithms and benchmark problems, faster execution of Java compared with Python) and jMetalPy (better support for data analysis and visualization). We detail an example of combining both frameworks in the next section.

\subsection{Experimental Case Study}

Let us consider the following case study. We are interested in assessing the performance of five metaheuristics (GDE3, MOCell, MOEA/D, NSGA-II, and SMPSO) when solving the ZDT suite of continuous problems (ZDT1-4, ZDT6). The quality indicators to calculate are set to the additive Epsilon (EP), Spread (SPREAD), and Hypervolume (HV), which give a measure of convergence, diversity and both properties, respectively. The number of independent runs for every algorithm is set to 25.

We configure an experiment with this information in jMetal and, after running the algorithms and applying the quality indicators, the summary file is obtained. Then, by giving this file as input to the jMetalPy statistical analysis module, we obtain a set of \LaTeX{} files and figures in an output directory as well as information displayed in the screen. We analyze next the obtained results.
\begin{table}[!htp]
  \caption{Median and Interquartile Range of the EP quality indicator.}
  \label{table:ep}
  \centering
  \resizebox{\columnwidth}{!}{
  \begin{tabular}{c|ccccc}
      & \textbf{NSGAII} & \textbf{SMPSO} & \textbf{MOEAD} & \textbf{GDE3} & \textbf{MOCell} \\\hline
      \textbf{ZDT1} & $1.29e-02_{2.69e-03} $ & $ \cellcolor{gray95} 5.59e-03_{2.64e-04} $ & $ 2.50e-02_{9.32e-03} $ & $ 1.31e-02_{2.96e-03} $ & $ \cellcolor{gray25} 6.26e-03_{2.44e-04}$ \\
      \textbf{ZDT2} & $1.33e-02_{2.63e-03} $ & $ \cellcolor{gray95} 5.47e-03_{2.82e-04} $ & $ 4.78e-02_{2.27e-02} $ & $ 1.25e-02_{3.16e-03} $ & $ \cellcolor{gray25} 5.72e-03_{2.72e-04}$ \\
      \textbf{ZDT3} & $7.94e-03_{2.27e-03} $ & $ \cellcolor{gray25} 5.23e-03_{1.22e-03} $ & $ 1.02e-01_{2.68e-02} $ & $ 7.13e-03_{1.39e-03} $ & $ \cellcolor{gray95} 5.19e-03_{1.24e-03}$ \\
      \textbf{ZDT4} & $1.42e-02_{2.43e-03} $ & $ \cellcolor{gray95} 6.12e-03_{4.06e-04} $ & $ 4.05e-01_{4.32e-01} $ & $ 4.08e+00_{8.64e-01} $ & $ \cellcolor{gray25} 9.07e-03_{2.65e-03}$ \\
      \textbf{ZDT6} & $1.97e-02_{3.62e-03} $ & $ \cellcolor{gray95} 6.79e-03_{2.85e-04} $ & $ \cellcolor{gray25} 7.73e-03_{1.23e-04} $ & $ 1.73e-02_{3.73e-03} $ & $ 8.43e-03_{8.69e-04}$ \\
  \end{tabular}
}
\end{table}

\begin{table}[!htp]
  \caption{Median and Interquartile Range of the SPREAD quality indicator.}
  \label{table:spread}
  \centering
  \resizebox{\columnwidth}{!}{
  \begin{tabular}{c|ccccc}
      & \textbf{NSGAII} & \textbf{SMPSO} & \textbf{MOEAD} & \textbf{GDE3} & \textbf{MOCell} \\\hline
      \textbf{ZDT1} & $3.45e-01_{2.80e-02} $ & $ \cellcolor{gray95} 6.92e-02_{1.95e-02} $ & $ 3.56e-01_{5.41e-02} $ & $ 3.33e-01_{3.05e-02} $ & $ \cellcolor{gray25} 7.17e-02_{1.44e-02}$ \\
      \textbf{ZDT2} & $3.63e-01_{3.84e-02} $ & $ \cellcolor{gray95} 7.19e-02_{1.31e-02} $ & $ 2.97e-01_{9.69e-02} $ & $ 3.33e-01_{3.95e-02} $ & $ \cellcolor{gray25} 8.50e-02_{2.30e-02}$ \\
      \textbf{ZDT3} & $7.47e-01_{1.50e-02} $ & $ \cellcolor{gray25} 7.10e-01_{1.07e-02} $ & $ 9.96e-01_{4.02e-02} $ & $ 7.34e-01_{1.26e-02} $ & $ \cellcolor{gray95} 7.04e-01_{5.88e-03}$ \\
      \textbf{ZDT4} & $3.57e-01_{2.93e-02} $ & $ \cellcolor{gray95} 9.04e-02_{1.26e-02} $ & $ 9.53e-01_{1.32e-01} $ & $ 8.92e-01_{6.10e-02} $ & $ \cellcolor{gray25} 1.20e-01_{3.50e-02}$ \\
      \textbf{ZDT6} & $4.71e-01_{2.76e-02} $ & $ \cellcolor{gray95} 2.49e-01_{1.06e-02} $ & $ 2.91e-01_{6.55e-04} $ & $ 6.73e-01_{3.90e-02} $ & $ \cellcolor{gray25} 2.68e-01_{1.22e-02}$ \\
  \end{tabular}
}
\end{table}

\begin{table}[!htp]
    \caption{Median and Interquartile Range of the HV quality indicator.}
    \label{table:hv}
    \centering
  \resizebox{\columnwidth}{!}{
    \begin{tabular}{c|ccccc}
      & \textbf{NSGAII} & \textbf{SMPSO} & \textbf{MOEAD} & \textbf{GDE3} & \textbf{MOCell} \\\hline
      \textbf{ZDT1} & $6.59e-01_{3.73e-04} $ & $ \cellcolor{gray95} 6.62e-01_{1.09e-04} $ & $ 6.42e-01_{5.71e-03} $ & $ 6.61e-01_{1.89e-04} $ & $ \cellcolor{gray25} 6.61e-01_{1.72e-04}$ \\
      \textbf{ZDT2} & $3.26e-01_{3.39e-04} $ & $ \cellcolor{gray95} 3.29e-01_{1.18e-04} $ & $ 3.12e-01_{6.94e-03} $ & $ 3.27e-01_{2.89e-04} $ & $ \cellcolor{gray25} 3.28e-01_{1.97e-04}$ \\
      \textbf{ZDT3} & $5.15e-01_{2.53e-04} $ & $ \cellcolor{gray25} 5.15e-01_{6.44e-04} $ & $ 4.41e-01_{2.99e-02} $ & $ \cellcolor{gray95} 5.15e-01_{1.28e-04} $ & $ 5.15e-01_{3.51e-04}$ \\
      \textbf{ZDT4} & $6.57e-01_{3.38e-03} $ & $ \cellcolor{gray95} 6.61e-01_{2.10e-04} $ & $ 2.76e-01_{2.33e-01} $ & $ 0.00e+00_{0.00e+00} $ & $ \cellcolor{gray25} 6.58e-01_{1.87e-03}$ \\
      \textbf{ZDT6} & $3.88e-01_{1.63e-03} $ & $ \cellcolor{gray95} 4.00e-01_{9.21e-05} $ & $ \cellcolor{gray25} 4.00e-01_{2.92e-06} $ & $ 3.97e-01_{5.83e-04} $ & $ 3.97e-01_{1.20e-03}$ \\
  \end{tabular}
}
\end{table}


Tables~\ref{table:ep}, \ref{table:spread}, and~\ref{table:hv} show the median and interquartile range of the three selected quality indicators. To facilitate the analysis of the tables, some cells have a grey background. Two grey levels are used, dark and light, to highlight the algorithms yielding the best and second best indicator values, respectively (note that this is automatically performed by jMetalPy). From the tables, we can observe that SMPSO is the overall best performing algorithm, achieving the best indicator values in four problems and one second best value.

Nevertheless, it is well known that taking into account only median values for algorithm ranking does not ensure that their differences are statistically significant. Statistical rankings are also needed if we intend to rank the algorithm performance considering all the problems globally. Finally, in studies involving a large number of problems (we have used only five for simplicity), the visual analysis of the medians can be very complicated, so statistical diagrams gathering all the information are needed. This is the reason why jMetalPy can also generate a second set of \LaTeX{} tables compiling, in a visually intuitive fashion, the result of non-parametric null hypothesis significance tests run over a certain quality indicator for all algorithms. Tables \ref{table:wep}, \ref{table:wspread} and \ref{table:whv} are three examples of these tables computed by using the Wilcoxon rank sum test between every pair of algorithms (at the 5\% level of significance) for the EP, SPREAD and HV indicators, respectively. In each cell, results for each of the 5 datasets are represented by using three symbols: -- if there is not statistical significance between the algorithms represented by the row and column of the cell; $\triangledown$ if the approach labeling the column is statistically better than the algorithm in the row; and $\blacktriangle$ if the algorithm in the row significantly outperforms the approach in the column. 
\begin{table}[!htp]
  \caption{Wilcoxon values of the EP quality indicator (ZDT1, ZDT2, ZDT3, ZDT4, ZDT6).}
  \label{table:wep}
  \centering
  \resizebox{\columnwidth}{!}{
  \begin{tabular}{c|cccc}
      & \textbf{SMPSO} & \textbf{MOEAD} & \textbf{GDE3} & \textbf{MOCell} \\\hline
      \textbf{NSGAII} & $\blacktriangle\ \blacktriangle\ \blacktriangle\ \blacktriangle\ \blacktriangle\  $ & $ \triangledown\ \triangledown\ \triangledown\ \triangledown\ \blacktriangle\  $ & $ \text{--}\ \text{--}\ \blacktriangle\ \triangledown\ \blacktriangle\  $ & $ \blacktriangle\ \blacktriangle\ \blacktriangle\ \blacktriangle\ \blacktriangle\ $ \\
      \textbf{SMPSO} & $ $ & $ \triangledown\ \triangledown\ \triangledown\ \triangledown\ \triangledown\  $ & $ \triangledown\ \triangledown\ \triangledown\ \triangledown\ \triangledown\  $ & $ \triangledown\ \triangledown\ \text{--}\ \triangledown\ \triangledown\ $ \\
      \textbf{MOEAD} & $ $ & $  $ & $ \blacktriangle\ \blacktriangle\ \blacktriangle\ \triangledown\ \triangledown\  $ & $ \blacktriangle\ \blacktriangle\ \blacktriangle\ \blacktriangle\ \triangledown\ $ \\
      \textbf{GDE3} & $ $ & $  $ & $  $ & $ \blacktriangle\ \blacktriangle\ \blacktriangle\ \blacktriangle\ \blacktriangle\ $ \\
  \end{tabular}
}
\end{table}

\begin{table}[!htp]
  \caption{Wilcoxon values of the SPREAD quality indicator (ZDT1, ZDT2, ZDT3, ZDT4, ZDT6).}
  \label{table:wspread}
  \centering
  \resizebox{\columnwidth}{!}{
  \begin{tabular}{c|cccc}
      & \textbf{SMPSO} & \textbf{MOEAD} & \textbf{GDE3} & \textbf{MOCell} \\\hline
      \textbf{NSGAII} & $\blacktriangle\ \blacktriangle\ \blacktriangle\ \blacktriangle\ \blacktriangle\  $ & $ \triangledown\ \blacktriangle\ \triangledown\ \triangledown\ \blacktriangle\  $ & $ \text{--}\ \blacktriangle\ \blacktriangle\ \triangledown\ \triangledown\  $ & $ \blacktriangle\ \blacktriangle\ \blacktriangle\ \blacktriangle\ \blacktriangle\ $ \\
      \textbf{SMPSO} & $ $ & $ \triangledown\ \triangledown\ \triangledown\ \triangledown\ \triangledown\  $ & $ \triangledown\ \triangledown\ \triangledown\ \triangledown\ \triangledown\  $ & $ \text{--}\ \triangledown\ \blacktriangle\ \triangledown\ \triangledown\ $ \\
      \textbf{MOEAD} & $ $ & $  $ & $ \blacktriangle\ \triangledown\ \blacktriangle\ \blacktriangle\ \triangledown\  $ & $ \blacktriangle\ \blacktriangle\ \blacktriangle\ \blacktriangle\ \blacktriangle\ $ \\
      \textbf{GDE3} & $ $ & $  $ & $  $ & $ \blacktriangle\ \blacktriangle\ \blacktriangle\ \blacktriangle\ \blacktriangle\ $ \\
  \end{tabular}
}
\end{table}

\begin{table}[!htp]
  \caption{Wilcoxon values of the HV quality indicator (ZDT1, ZDT2, ZDT3, ZDT4, ZDT6).}
  \label{table:whv}
  \centering
    \resizebox{\columnwidth}{!}{
  \begin{tabular}{c|cccc}
      & \textbf{SMPSO} & \textbf{MOEAD} & \textbf{GDE3} & \textbf{MOCell} \\\hline
      \textbf{NSGAII} & $\triangledown\ \triangledown\ \triangledown\ \triangledown\ \triangledown\  $ & $ \blacktriangle\ \blacktriangle\ \blacktriangle\ \blacktriangle\ \triangledown\  $ & $ \triangledown\ \triangledown\ \triangledown\ \blacktriangle\ \triangledown\  $ & $ \triangledown\ \triangledown\ \text{--}\ \triangledown\ \triangledown\ $ \\
      \textbf{SMPSO} & $ $ & $ \blacktriangle\ \blacktriangle\ \blacktriangle\ \blacktriangle\ \blacktriangle\  $ & $ \blacktriangle\ \blacktriangle\ \text{--}\ \blacktriangle\ \blacktriangle\  $ & $ \blacktriangle\ \blacktriangle\ \blacktriangle\ \blacktriangle\ \blacktriangle\ $ \\
      \textbf{MOEAD} & $ $ & $  $ & $ \triangledown\ \triangledown\ \triangledown\ \blacktriangle\ \blacktriangle\  $ & $ \triangledown\ \triangledown\ \triangledown\ \triangledown\ \blacktriangle\ $ \\
      \textbf{GDE3} & $ $ & $  $ & $  $ & $ \triangledown\ \triangledown\ \blacktriangle\ \triangledown\ \text{--}\ $ \\
  \end{tabular}
}
\end{table}

\begin{figure}[h!]
    \centering
    \includegraphics[width=0.8\columnwidth]{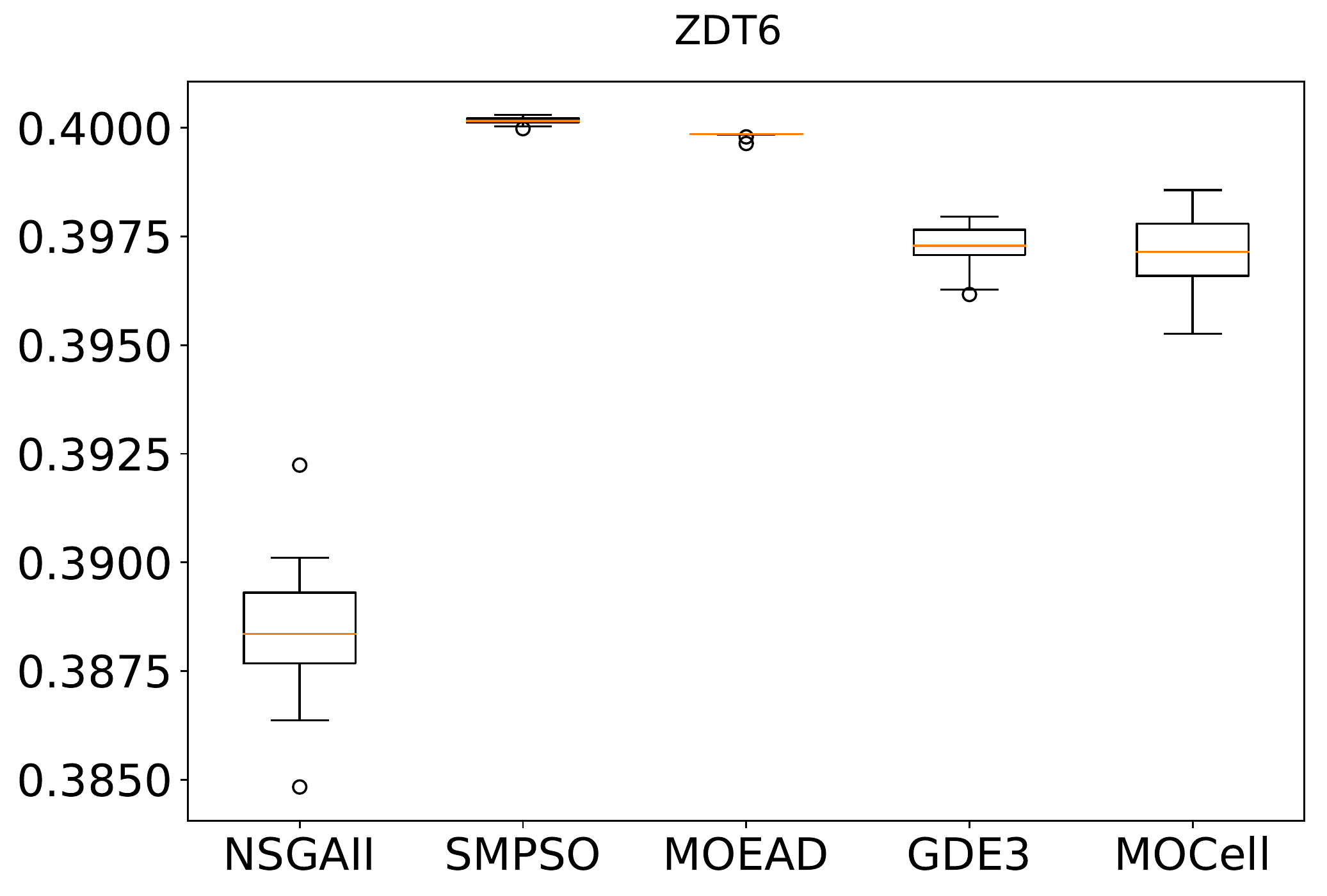}
    \caption{Boxplot diagram of the HV indicator for the ZDT6 problem.}
    \label{figure:boxplot-hv}
\end{figure}

The conclusions drawn from the above tables can be buttressed by inspecting the distribution of the quality indicator values obtained by the algorithms. Figure~\ref{figure:boxplot-hv} shows boxplots obtained with jMetalPy by means of the Hypervolume values when solving the ZDT6 problem. Whenever the boxes do not overlap with each other we can state that there should be statistical confidence that the performance gaps are relevant (e.g., between SMPSO and the rest of algorithms), but when they do (as we can see with GDE3 and MOCell) we cannot discern which algorithm performs best.

\begin{figure}[!ht]
    \centering
    \includegraphics[width=0.95\columnwidth]{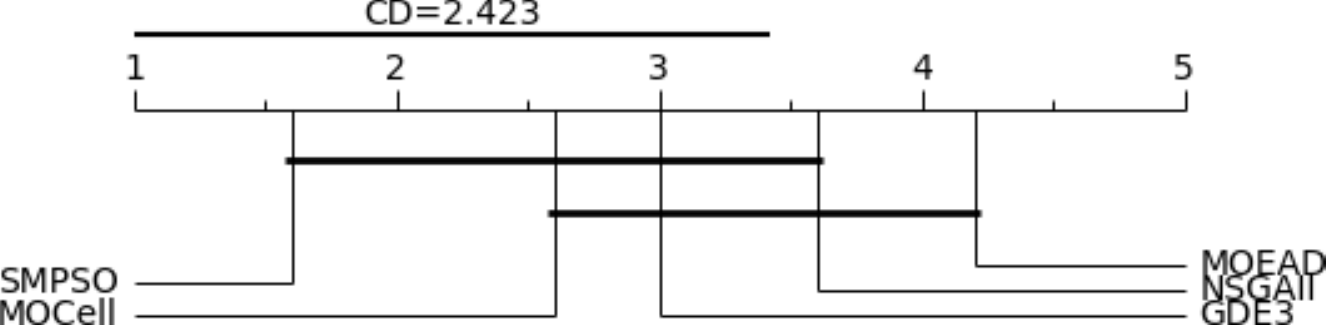}
    \caption{CD plot of the HV indicator.}
    \label{figure:cdplot}
\end{figure}

\begin{figure}[!ht]
    \centering
    \includegraphics[width=80mm]{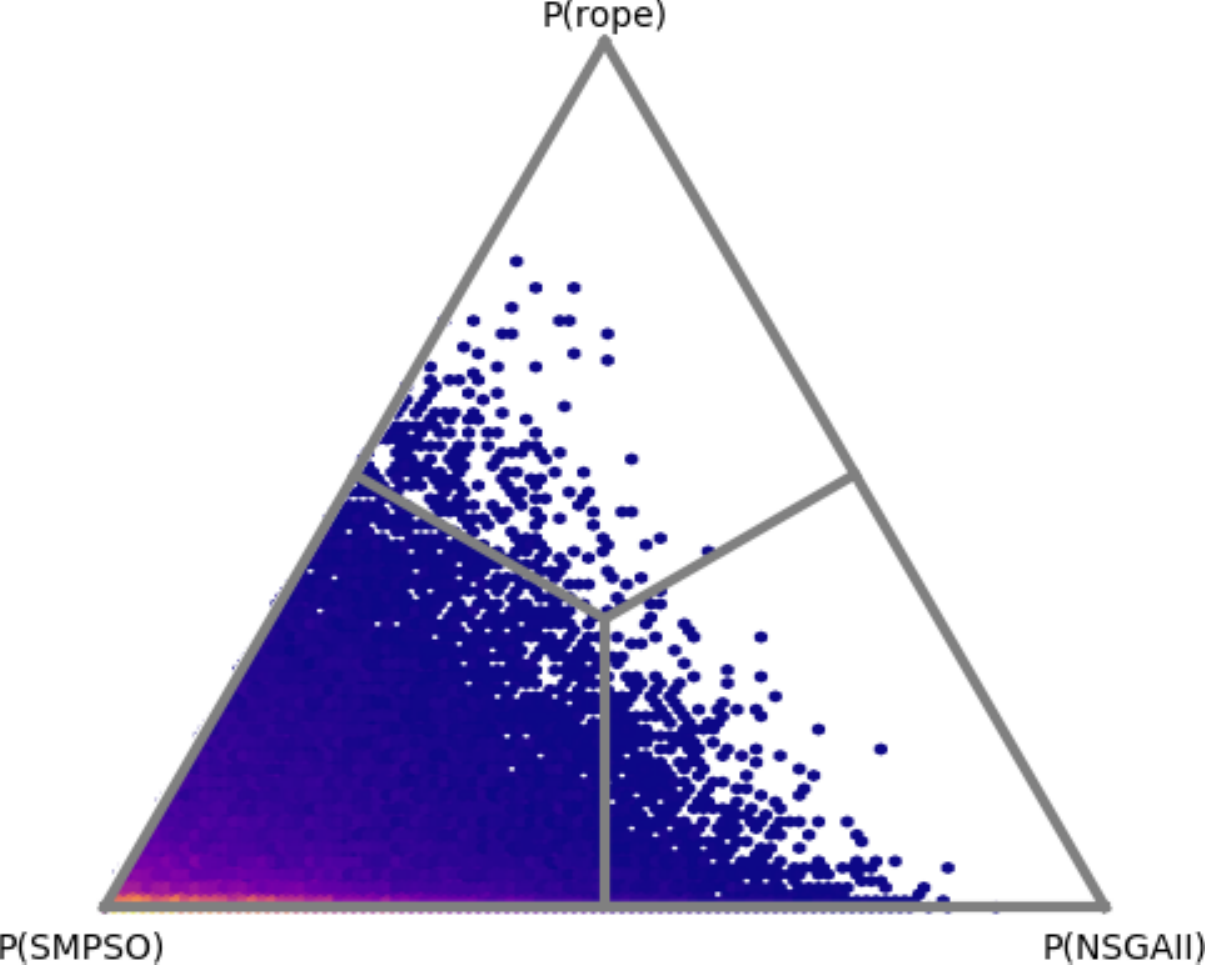}
    \caption{Posterior plot of the HV indicator using a Bayesian sign test.}
    \label{figure:postplot}
\end{figure}

The boxplots and tables described heretofore allow observing the dispersion of the results, as well as the presence of outliers, but they do not allow to get a global vision of the performance of the algorithms in all the problems. This motivates the incorporation of principled methods for comparing multiple techniques over different problem instances, such as those proposed by Demsar \cite{demvsar2006statistical} in the context of classification problems and machine learning models. As anticipated previously, our developed framework includes Critical Distance (CD) plots (Figure~\ref{figure:cdplot}, computed for the HV indicator) and Posterior plots (Figure~\ref{figure:postplot}, again for the HV indicator). The former plot is used to depict the average ranking of the algorithms computed over the considered problems, so that the chart connects with a bold line those algorithms whose difference in ranks is less than the so-called critical distance. This critical distance is a function of the number of problems, the number of techniques under comparison, and a critical value that results from a Studentized range statistic and a specified confidence level. As shown in Figure \ref{figure:cdplot}, SMPSO, MOCell, GDE3 and NSGA-II are reported to perform statistically equivalently, which clashes with the conclusions of the previously discussed table set due to the relatively higher strictness of the statistic from which the critical distance is computed. A higher number of problems would be required to declare statistical significance under this comparison approach.

Finally, we end up our experimental use case by showing the Posterior plot that allows comparing pair of algorithms by using Bayesian sign test (Figure \ref{figure:postplot}). When relying on Bayesian hypothesis testing we can directly evaluate the posterior probability of the hypotheses from the available quality indicator values, which enables a more intuitive understanding of the comparative performance of the considered algorithms. Furthermore, a region of practical equivalence (also denoted \emph{rope}) can be defined to account for ties between the considered multi-objective solvers. The plot in Figure \ref{figure:postplot} is in essence a barycentric projection of the posterior probabilities: the region at the bottom-right of the chart, for instance, delimits the area where:
\begin{equation}
\theta_r\geq\max(\theta_e,\theta_l),
\end{equation}
with $\theta_r=P(z>r)$, $\theta_e=P(-r\leq z \leq r)$, $\theta_l=P(z<-r)$, and $z$ denoting the difference between the indicator values of the algorithm on the right and the left (in that order). Based on this notation, the figure exposes, in our case and for $r=0.002$, than in most cases $z=HV(\mbox{NSGA-II})-HV(\mbox{SMPSO})$ fulfills $\theta_l\geq\max(\theta_e,\theta_r)$, i.e. it is more probable, on average, than the HV values of SMPSO are higher than those of NSGA-II. Particularly these probabilities can be estimated by counting the number of points that fall in every one of the three regions, from which we conclude that in this use case 1) SMPSO is practically better than NSGA-II with probability 0.918; 2) both algorithms perform equivalently with probability 0.021; and 3) NSGA-II is superior than SMPSO with probability 0.061.

\section{Conclusions and Future Work} \label{sec:conclusions}

In this paper we have presented jMetalPy, a Python-based framework for multi-objective optimization with metaheuristics. It is released under the MIT license and made freely available for the community in GitHub. We have detailed its core architecture and described the implementation of NSGA-II and some of its variants as illustrative examples of how to operate with this framework. jMetalPy provides support for dynamic optimization, parallelism, and decision making. Other salient features involves visualization (static, streaming, and interactive graphics) for multi- and many-objective problems, and a large set of statistical tests for performance assessment. It is worth noting that jMetalPy is still a young research project, which is intended to be useful for the research community interested in multi-objective optimization with metaheuristics. Thus, it is expected to evolve quickly, incorporating new algorithms and problems by both the development team and by external contributors. 

Specific lines of future work include evaluating the performance of parallel and distributed metaheuristics in clusters, as well as applying them to solve real-world problems. 

\section*{Acknowledgements}
This work has been partially funded by Grants TIN2017-86049-R (Spanish Ministry of Education and Science). Jos\'e Garc\'ia-Nieto is the recipient of a Post-Doctoral fellowship of ``Captaci\'on de Talento para la Investigaci\'on'' Plan Propio at Universidad de M\'alaga. Javier Del Ser and Izaskun Oregui receive funding support from the Basque Government through the EMAITEK Program.

\section*{Bibliography}

\bibliographystyle{plain}
\bibliography{refs}

\begin{thebibliography}{10}
\expandafter\ifx\csname url\endcsname\relax
  \def\url#1{\texttt{#1}}\fi
\expandafter\ifx\csname urlprefix\endcsname\relax\def\urlprefix{URL }\fi
\expandafter\ifx\csname href\endcsname\relax
  \def\href#1#2{#2} \def\path#1{#1}\fi

\bibitem{Coello2006}
C.~A.~C. Coello, G.~B. Lamont, D.~A.~V. Veldhuizen, Evolutionary Algorithms for
  Solving Multi-Objective Problems (Genetic and Evolutionary Computation),
  Springer-Verlag, Berlin, Heidelberg, 2006.

\bibitem{Deb01}
K.~Deb, Multi-Objective Optimization Using Evolutionary Algorithms, John Wiley
  \& Sons, 2001.

\bibitem{Weise09}
T.~Weise, M.~Zapf, R.~Chiong, A.~J. Nebro, {Why Is Optimization Difficult?},
  in: R.~Chiong (Ed.), Nature-Inspired Algorithms for Optimisation, Springer,
  Berlin, 2009, pp. 1--50, iSBN 978-3-642-00266-3.

\bibitem{BR2003}
C.~Blum, A.~Roli, {Metaheuristics in combinatorial optimization: Overview and
  conceptual comparison}, ACM Computing Surveys 35~(3) (2003) 268--308.

\bibitem{Durillo2011}
J.~J. Durillo, A.~J. Nebro, jmetal: A java framework for multi-objective
  optimization, Advances in Engineering Software 42~(10) (2011) 760--771.

\bibitem{NDV15}
A.~Nebro, J.~J. Durillo, M.~Vergne, Redesigning the jmetal multi-objective
  optimization framework, in: Proceedings of the Companion Publication of the
  2015 Annual Conference on Genetic and Evolutionary Computation, GECCO
  Companion '15, ACM, 2015, pp. 1093--1100.

\bibitem{Numpy}
T.~Oliphant, {NumPy}: A guide to {NumPy}, \url{http://www.numpy.org/}, [Online;
  accessed 02-08-2019] (2006).

\bibitem{Scipy}
E.~Jones, T.~Oliphant, P.~Peterson, et~al.,
  \href{http://www.scipy.org/}{{SciPy}: Open source scientific tools for
  {Python}}, [Online; accessed 02-08-2019] (2001--).
\newline\urlprefix\url{http://www.scipy.org/}

\bibitem{scikit-learn}
F.~Pedregosa, G.~Varoquaux, A.~Gramfort, V.~Michel, B.~Thirion, O.~Grisel,
  M.~Blondel, P.~Prettenhofer, R.~Weiss, V.~Dubourg, J.~Vanderplas, A.~Passos,
  D.~Cournapeau, M.~Brucher, M.~Perrot, E.~Duchesnay, Scikit-learn: Machine
  learning in {P}ython, Journal of Machine Learning Research 12 (2011)
  2825--2830.

\bibitem{Hunter:2007}
J.~D. Hunter, Matplotlib: A 2d graphics environment, Computing In Science \&
  Engineering 9~(3) (2007) 90--95.
\newblock \href {https://doi.org/10.1109/MCSE.2007.55}
  {\path{doi:10.1109/MCSE.2007.55}}.

\bibitem{holo}
J.-L. R~Stevens, P.~Rudiger, J.~A~Bednar, Holoviews: Building complex
  visualizations easily for reproducible science, 2015.
\newblock \href {https://doi.org/10.25080/Majora-7b98e3ed-00a}
  {\path{doi:10.25080/Majora-7b98e3ed-00a}}.

\bibitem{plotly}
P.~T. Inc., \href{https://plot.ly}{Collaborative data science} (2015).
\newline\urlprefix\url{https://plot.ly}

\bibitem{Dask}
{Dask Development Team}, \href{https://dask.org}{Dask: Library for dynamic task
  scheduling} (2016).
\newline\urlprefix\url{https://dask.org}

\bibitem{salloum2016big}
S.~Salloum, R.~Dautov, X.~Chen, P.~X. Peng, J.~Z. Huang, Big data analytics on
  apache spark, International Journal of Data Science and Analytics 1~(3-4)
  (2016) 145--164.

\bibitem{DPA02}
K.~Deb, A.~Pratap, S.~Agarwal, T.~Meyarivan, {A Fast and Elitist Multiobjective
  Genetic Algorithm: {NSGA-II}}, IEEE Trans. Evol. Comput. 6~(2) (2002)
  182--197.

\bibitem{Kukkonen2005}
S.~Kukkonen, J.~Lampinen, {GDE3: the third evolution step of generalized
  differential evolution}, in: Evolutionary Computation, 2005. The 2005 IEEE
  Congress on, Vol.~1, 2005, pp. 443--450.
\newblock \href {https://doi.org/10.1109/CEC.2005.1554717}
  {\path{doi:10.1109/CEC.2005.1554717}}.

\bibitem{Nebro2009}
A.~J. Nebro, J.~J. Durillo, J.~Garcia-Nieto, C.~A. Coello~Coello, F.~Luna,
  E.~Alba, {SMPSO: A new PSO-based metaheuristic for multi-objective
  optimization}, in: {IEEE Symposium on Computational Intelligence in
  Multi-Criteria Decision-Making}, 2009, pp. 66--73.
\newblock \href {https://doi.org/10.1109/MCDM.2009.4938830}
  {\path{doi:10.1109/MCDM.2009.4938830}}.

\bibitem{omopso2004}
C.~A.~C. Coello, G.~T. Pulido, M.~S. Lechuga, Handling multiple objectives with
  particle swarm optimization, IEEE Transactions on Evolutionary Computation
  8~(3) (2004) 256--279.
\newblock \href {https://doi.org/10.1109/TEVC.2004.826067}
  {\path{doi:10.1109/TEVC.2004.826067}}.

\bibitem{Qingfu2007}
Q.~Zhang, H.~Li, {MOEA/D: A Multiobjective Evolutionary Algorithm Based on
  Decomposition}, IEEE T. Evolut. Comput. 11~(6) (2007) 712--731.
\newblock \href {https://doi.org/10.1109/TEVC.2007.892759}
  {\path{doi:10.1109/TEVC.2007.892759}}.

\bibitem{LZ09}
H.~Li, Q.~Zhang, {Multiobjective Optimization Problems With Complicated Pareto
  Sets, MOEA/D and NSGA-II}, IEEE Transactions on Evolutionary Computation
  13~(2) (2009) 229--242.

\bibitem{FDA04}
M.~Farina, K.~Deb, P.~Amato, Dynamic multiobjective optimization problems: test
  cases, approximations, and applications, IEEE Transactions on Evolutionry
  Computation 8~(5) (2004) 425--442.
\newblock \href {https://doi.org/10.1109/TEVC.2004.831456}
  {\path{doi:10.1109/TEVC.2004.831456}}.

\bibitem{smpsorp2018}
A.~J. Nebro, J.~J. Durillo, J.~Garc{\'i}a-Nieto, C.~Barba-Gonz{\'a}lez,
  J.~Del~Ser, C.~A. Coello~Coello, A.~Ben{\'i}tez-Hidalgo, J.~F. Aldana-Montes,
  Extending the speed-constrained multi-objective pso (smpso) with reference
  point based preference articulation, in: A.~Auger, C.~M. Fonseca,
  N.~Louren{\c{c}}o, P.~Machado, L.~Paquete, D.~Whitley (Eds.), Parallel
  Problem Solving from Nature -- PPSN XV, Springer International Publishing,
  Cham, 2018, pp. 298--310.

\bibitem{ZT99}
E.~Zitzler, L.~Thiele, Multiobjective evolutionary algorithms: a comparative
  case study and the strength pareto approach, IEEE Transactions on
  Evolutionary Computation 3~(4) (1999) 257--271.
\newblock \href {https://doi.org/10.1109/4235.797969}
  {\path{doi:10.1109/4235.797969}}.

\bibitem{ZTL+03}
E.~Zitzler, L.~Thiele, M.~Laumanns, C.~M. Fonseca, V.~G. da~Fonseca,
  Performance assessment of multiobjective optimizers: an analysis and review,
  IEEE Transactions on Evolutionary Computation 7~(2) (2003) 117--132.
\newblock \href {https://doi.org/10.1109/TEVC.2003.810758}
  {\path{doi:10.1109/TEVC.2003.810758}}.

\bibitem{CR04}
C.~A. Coello~Coello, M.~Reyes~Sierra, A study of the parallelization of a
  coevolutionary multi-objective evolutionary algorithm, in: R.~Monroy,
  G.~Arroyo-Figueroa, L.~E. Sucar, H.~Sossa (Eds.), MICAI 2004: Advances in
  Artificial Intelligence, Springer Berlin Heidelberg, Berlin, Heidelberg,
  2004, pp. 688--697.

\bibitem{Zaharia+2010}
M.~Zaharia, M.~Chowdhury, M.~J. Franklin, S.~Shenker, I.~Stoica, Spark: Cluster
  computing with working sets, in: Proceedings of the 2Nd USENIX Conference on
  Hot Topics in Cloud Computing, HotCloud'10, USENIX Association, 2010, pp.
  10--10.

\bibitem{DEAP_JMLR2012}
F.-A. Fortin, F.-M. {De Rainville}, M.-A. Gardner, M.~Parizeau, C.~Gagn\'e,
  {DEAP}: Evolutionary algorithms made easy, Journal of Machine Learning
  Research 13 (2012) 2171--2175.

\bibitem{geatpy}
G.~core team, {Geatpy - The Genetic and Evolutionary Algorithm Toolbox for
  Python}, \url{http://www.geatpy.com}, [Online; accessed: 01-21-2019] (2018).

\bibitem{inspyred}
A.~Garrett, {inspyred (Version 1.0.1) [software]. Inspired Intelligence},
  \url{https://github.com/aarongarrett/inspyred}, [Online; accessed:
  01-08-2019] (2012).

\bibitem{pagmo}
F.~Biscani, D.~Izzo, C.~H. Yam, A global optimisation toolbox for massively
  parallel engineering optimisation, arXiv:1004.3824 [cs.DC]
  \url{https://esa.github.io/pagmo2/index.html}, [Online; accessed: 01-18-2019]
  (2010).

\bibitem{Platypus}
D.~Hadka, {Platypus. A Free and Open Source Python Library for Multiobjective
  Optimization}, \url{https://github.com/Project-Platypus/Platypus}, [Online;
  accessed: 01-08-2019] (2015).

\bibitem{blank19}
J.~Blank, pymoo - multi-objective optimization framework,
  \url{https://github.com/msu-coinlab/pymoo} (2019).

\bibitem{Luke2017}
S.~Luke, \href{http://doi.acm.org/10.1145/3067695.3082467}{Ecj then and now},
  in: Proceedings of the Genetic and Evolutionary Computation Conference
  Companion, GECCO '17, ACM, New York, NY, USA, 2017, pp. 1223--1230.
\newblock \href {https://doi.org/10.1145/3067695.3082467}
  {\path{doi:10.1145/3067695.3082467}}.
\newline\urlprefix\url{http://doi.acm.org/10.1145/3067695.3082467}

\bibitem{WZ97}
J.~Wakunda, A.~Zell, Eva: a tool for optimization with evolutionary algorithms,
  in: EUROMICRO 97. Proceedings of the 23rd EUROMICRO Conference: New Frontiers
  of Information Technology (Cat. No.97TB100167), 1997, pp. 644--651.
\newblock \href {https://doi.org/10.1109/EURMIC.1997.617395}
  {\path{doi:10.1109/EURMIC.1997.617395}}.

\bibitem{jclec15}
A.~Ram{\'{\i}}rez, J.~R. Romero, S.~Ventura,
  \href{https://doi.org/10.1145/2739482.2768461}{An extensible jclec-based
  solution for the implementation of multi-objective evolutionary algorithms},
  in: Genetic and Evolutionary Computation Conference, {GECCO} 2015, Madrid,
  Spain, July 11-15, 2015, Companion Material Proceedings, 2015, pp.
  1085--1092.
\newblock \href {https://doi.org/10.1145/2739482.2768461}
  {\path{doi:10.1145/2739482.2768461}}.
\newline\urlprefix\url{https://doi.org/10.1145/2739482.2768461}

\bibitem{MOEAFramework}
D.~Hadka, Moea framework: A free and open source java framework for
  multiobjective optimization, \url{http://moeaframework.org/}, [Online;
  accessed 02-08-2019] (2017).

\bibitem{opt4jpaper}
M.~Lukasiewycz, M.~Gla{\ss}, F.~Reimann, J.~Teich, {Opt4J - A Modular Framework
  for Meta-heuristic Optimization}, in: Proceedings of the Genetic and
  Evolutionary Computing Conference (GECCO 2011), Dublin, Ireland, 2011, pp.
  1723--1730.

\bibitem{Liefooghe10a}
A.~Liefooghe, L.~Jourdan, T.~Legrand, J.~Humeau, E.-G. Talbi, {ParadisEO-MOEO:
  A Software Framework for Evolutionary Multi-Objective Optimization}, in:
  C.~A. {Coello Coello}, C.~Dhaenens, L.~Jourdan (Eds.), Advances in
  Multi-Objective Nature Inspired Computing, Springer, Studies in Computational
  Intelligence, Vol. 272, Berlin, Germany, 2010, Ch.~5, pp. 87--117, iSBN
  978-3-642-11217-1.

\bibitem{Bleuler02}
S.~Bleuler, M.~Laumanns, L.~Thiele, E.~Zitzler, {PISA --- A Platform and
  Programming Language Independent Interface for Search Algorithms}, TIK Report
  154, {Computer Engineering and Networks Laboratory (TIK), ETH Zurich}
  (October 2002).

\bibitem{Platemo17}
Y.~Tian, R.~Cheng, X.~Zhang, Y.~Jin, Platemo: A matlab platform for
  evolutionary multi-objective optimization [educational forum], IEEE
  Computational Intelligence Magazine 12~(4) (2017) 73--87.
\newblock \href {https://doi.org/10.1109/MCI.2017.2742868}
  {\path{doi:10.1109/MCI.2017.2742868}}.

\bibitem{GHJ+94}
E.~Gamma, R.~Helm, R.~Johnson, J.~Vlissides, Design Patterns: Elements of
  Reusable Object-Oriented Software, 1st Edition, Addison-Wesley Professional,
  1994.

\bibitem{zdt2000a}
E.~Zitzler, K.~Deb, L.~Thiele, {Comparison of Multiobjective Evolutionary
  Algorithms: Empirical Results}, Evolutionary Computation 8~(2) (2000)
  173--195.

\bibitem{DNL08}
J.~J. Durillo, A.~J. Nebro, F.~Luna, E.~Alba, A study of master-slave
  approaches to parallelize nsga-ii, in: 2008 IEEE International Symposium on
  Parallel and Distributed Processing, 2008, pp. 1--8.
\newblock \href {https://doi.org/10.1109/IPDPS.2008.4536375}
  {\path{doi:10.1109/IPDPS.2008.4536375}}.

\bibitem{MOLINA2009685}
J.~Molina, L.~V. Santana, A.~G. Hernández-Díaz, C.~A.~C. Coello,
  R.~Caballero,
  \href{http://www.sciencedirect.com/science/article/pii/S0377221708005146}{g-dominance:
  Reference point based dominance for multiobjective metaheuristics}, European
  Journal of Operational Research 197~(2) (2009) 685 -- 692.
\newblock \href {https://doi.org/https://doi.org/10.1016/j.ejor.2008.07.015}
  {\path{doi:https://doi.org/10.1016/j.ejor.2008.07.015}}.
\newline\urlprefix\url{http://www.sciencedirect.com/science/article/pii/S0377221708005146}

\bibitem{BGN17}
C.~Barba-Gonzal{\'e}z, J.~Garc{\'i}a-Nieto, A.~J. Nebro, J.~F. Aldana-Montes,
  Multi-objective big data optimization with jmetal and spark, in:
  H.~Trautmann, G.~Rudolph, K.~Klamroth, O.~Sch{\"u}tze, M.~Wiecek, Y.~Jin,
  C.~Grimme (Eds.), Evolutionary Multi-Criterion Optimization, Springer
  International Publishing, Cham, 2017, pp. 16--30.

\bibitem{DTLZ01}
K.~Deb, L.~Thiele, M.~Laumanns, E.~Zitzler, Scalable {T}est {P}roblems for
  {E}volutionary {M}ultiobjective {O}ptimization, in: A.~Abraham, L.~Jain,
  R.~Goldberg (Eds.), Evolutionary {M}ultiobjective {O}ptimization.
  {T}heoretical {A}dvances and {A}pplications, Springer, 2001, pp. 105--145.

\bibitem{benavoli2017time}
A.~Benavoli, G.~Corani, J.~Dem{\v{s}}ar, M.~Zaffalon, Time for a change: a
  tutorial for comparing multiple classifiers through bayesian analysis, The
  Journal of Machine Learning Research 18~(1) (2017) 2653--2688.

\bibitem{demvsar2006statistical}
J.~Dem{\v{s}}ar, Statistical comparisons of classifiers over multiple data
  sets, Journal of Machine learning research 7~(Jan) (2006) 1--30.

\end{thebibliography}

\end{document}